\documentclass{article}

\usepackage{microtype}
\usepackage{graphicx}
\usepackage{booktabs} 
\usepackage{float}
\usepackage{lipsum} 
\usepackage{bm}
\usepackage{afterpage}
\usepackage{wrapfig}

\usepackage{graphicx}
\usepackage{subcaption}  

\usepackage{hyperref}

\usepackage{xcolor}

\definecolor{orangehighlight}{HTML}{FFB347}  
\definecolor{deepblue}{HTML}{0000CD}         
\definecolor{forestgreen}{HTML}{228B22}      
\definecolor{darkmagenta}{HTML}{8B008B}      

\usepackage{hyperref}

\usepackage[accepted]{icml2025}

\usepackage{amsmath}
\usepackage{amssymb}
\usepackage{pifont}
\usepackage{mathtools}
\usepackage{amsthm}

\usepackage{microtype}
\usepackage{graphicx}
\usepackage{booktabs} %
\usepackage{xcolor}
\usepackage{titletoc}
\usepackage{hyperref}
\usepackage{siunitx}
\usepackage[inline]{enumitem}

\usepackage{multirow}
\usepackage{array}
\usepackage{multicol}
\usepackage{tabularx}

\usepackage{adjustbox}
\usepackage{caption}
\usepackage{subcaption}

\definecolor{Red}{rgb}{0.768, 0.054, 0.054}
\definecolor{Blue}{rgb}{0.43, 0.65, 0.86}
\definecolor{Green}{rgb}{0,0.4,0.7}
\definecolor{hotpink}{rgb}{1.0, 0.41, 0.71}
\definecolor{brown}{rgb}{0.59, 0.29, 0.0}
\definecolor{darkpastelgreen}{rgb}{0.01, 0.75, 0.24}
\definecolor{celestialblue}{rgb}{0.29, 0.59, 0.82}
\definecolor{ceruleanblue}{rgb}{0.16, 0.32, 0.75}
\definecolor{goldenrod}{rgb}{0.85, 0.65, 0.13}
\definecolor{navyblue}{rgb}{0.0, 0.0, 0.5}
\definecolor{coolgrey}{rgb}{0.55, 0.57, 0.67}
\definecolor{darkseagreen}{rgb}{0.56, 0.74, 0.56}
\definecolor{darkturquoise}{rgb}{0.0, 0.81, 0.82}
\definecolor{berryred}{rgb}{0.79, 0.25, 0.14}
\definecolor{teagreen}{rgb}{0.81,0.94,0.75}
\definecolor{lightgrey}{rgb}{0.5,0.5,0.5}
\definecolor{purple}{rgb}{0.35,0.25,0.55}
\hypersetup{
    colorlinks=true,
    citecolor=teal,
    linkcolor=Red,
    urlcolor=Green,
}

\newcommand{\cmark}{\textcolor{teal}{\ding{51}}}%
\newcommand{\xmark}{\textcolor{red}{\ding{55}}}%

\usepackage{algorithm}
\usepackage{algorithmic}
\usepackage{colortbl}

\usepackage{siunitx}
\usepackage{standalone}

\newcommand{\pms}[1]{\ensuremath{{\scriptstyle\pm #1}}}

\usepackage{tcolorbox}

\usepackage[algo2e]{algorithm2e}
\newcolumntype{x}[1]{>{\centering\let\newline\\\arraybackslash\hspace{0pt}}p{#1}}

\usepackage[capitalize,noabbrev]{cleveref}

\icmltitlerunning{Robust Molecular Property Prediction via Densifying Scarce Labeled Data}

\begin{document}

\twocolumn[
\icmltitle{Robust Molecular Property Prediction via \\ Densifying Scarce Labeled Data}




\icmlsetsymbol{equal}{*}

\begin{icmlauthorlist}
\icmlauthor{Jina Kim}{equal,KA}
\icmlauthor{Jeffrey Willette}{equal,KA}
\icmlauthor{Bruno Andreis}{equal,KA}
\icmlauthor{Sung Ju Hwang}{KA,DA}
\end{icmlauthorlist}

\icmlaffiliation{KA}{Korea Advanced Institute of Science and Technology (KAIST), South Korea}
\icmlaffiliation{DA}{Deepauto.ai}

\icmlcorrespondingauthor{Jina Kim}{jinakim@kaist.ac.kr}
\icmlcorrespondingauthor{Jeffrey Willette}{jwillette@kaist.ac.kr}
\icmlcorrespondingauthor{Bruno Andreis}{andries@kaist.ac.kr}
\icmlcorrespondingauthor{Sung Ju Hwang}{sungju.hwang@kaist.ac.kr}

\icmlkeywords{Machine Learning, Generative AI, Biology, ICML}






\vskip 0.3in
]



\printAffiliationsAndNotice{\icmlEqualContribution} 

\begin{abstract}

A widely recognized limitation of molecular prediction models is their reliance on structures observed in the training data, resulting in poor generalization to out-of-distribution compounds. Yet in drug discovery, the compounds most critical for advancing research often lie beyond the training set, making the bias toward the training data particularly problematic. This mismatch introduces substantial covariate shift, under which standard deep learning models produce unstable and inaccurate predictions. Furthermore, the scarcity of labeled data—stemming from the onerous and costly nature of experimental validation—further exacerbates the difficulty of achieving reliable generalization. To address these limitations, we propose a novel bilevel optimization approach that leverages unlabeled data to interpolate between in-distribution (ID) and out-of-distribution (OOD) data, enabling the model to learn how to generalize beyond the training distribution. We demonstrate significant performance gains on challenging real-world datasets with substantial covariate shift, supported by t-SNE visualizations highlighting our interpolation method.
\end{abstract}

\begin{figure}[t]
\centering
\vspace{+0.0in}
\includegraphics[width=1.0\linewidth]{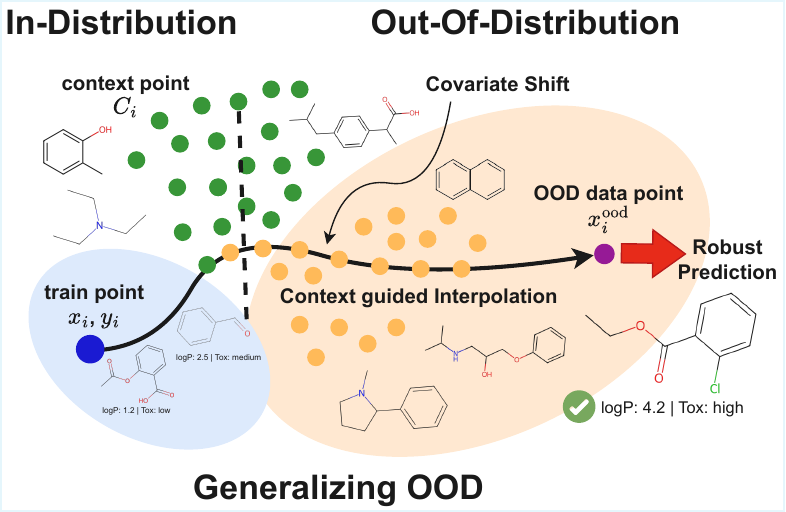}
\vspace{-1.8em}
\caption{\small \textbf{Concept.} We densify the train dataset using external unlabeled data (context point) for robust generalization across covariate shift. Notation details are provided in \cref{analysis}.}
\label{fig:concept_figure}
\vspace{-1.5em}
\end{figure}
\vspace{-2em}
\section{Introduction}
\label{intro}
Molecular property prediction plays a central role in drug discovery pipelines, enabling researchers to prioritize compounds for costly and time-consuming experimental validation. Accurate computational models have the potential to dramatically accelerate early-stage discovery by predicting critical attributes such as bioactivity, toxicity, and solubility before synthesis \citep{schneider2018automating, vamathevan2019applications}. However, building reliable predictive models generalizing to novel, unseen compounds remains a fundamental challenge.

Standard molecular property prediction models tend to rely heavily on patterns observed within the training distribution, resulting in poor generalization to out-of-distribution compounds \citep{qsavi, ovadia, koh}. In drug discovery, this limitation is particularly problematic, since the compounds most crucial for advancing research often lie far beyond the chemical spaces represented during training \citep{mood}. The resulting covariate shift introduces significant obstacles to reliable prediction, with models frequently producing unstable outputs when extrapolating to new regions of chemical space. Further compounding these challenges, experimental validation of molecular properties is both costly and resource-intensive, leading to a scarcity of labeled data and increasing reliance on computational exploration \citep{altae2017low}. Also, available labeled data is typically concentrated in narrow regions of chemical space, introducing bias that hampers generalization to unseen compounds \citep{qsavi}.

While vast collections of unlabeled molecular structures are readily available \citep{zinc, pubchem}, offering rich information about the structure of chemical space, existing methods often fail to fully exploit this resource to improve generalization \citep{qsavi}. Therefore, we propose a novel bilevel optimization method that leverages unlabeled data to densify the scarce train dataset and guide the model toward sensible behavior in unexplored regions of chemical space. Our code can be found at \url{https://github.com/JinA0218/drugood-densify}.

\section{Methodology}
\label{method}

\textbf{Preliminaries.}  We consider the problem of molecular property prediction under covariate shift. Given a small labeled dataset \(\mathcal{D}_\text{train} = \{(x_i, y_i)\}_{i=1}^n\) and abundant unlabeled molecules \(\mathcal{D}_\text{unlabeled} = \{x_j\}_{j=1}^m\), the goal is to learn a predictive model \(f: \mathcal{X} \to \mathcal{Y}\) that reliably generalizes to a distributionally shifted test set \(\mathcal{D}_\text{test}\).

\textbf{Scarce Data Densification with Unlabeled Data}
To address this, we propose a bilevel optimization framework that interpolates the training distribution  $\mathcal{D}_\text{train}$ with an exogenous distribution $\mathcal{D}_\text{unlabeled}$.
Our objective is to leverage the cheaper and more abundant distribution $\mathcal{D}_\text{unlabeled}$ to densify the scarce labeled distribution $\mathcal{D}_\text{train}$ in a way that encourages the model to generalize robustly under covariate shift, particularly in out-of-distribution scenarios where we have no label information and therefore high uncertainty.
For this, we utilize subsets of $\mathcal{D}_\text{unlabeled}$ as $\mathcal{D}_\text{context}$ and $\mathcal{D}_\text{mvalid}$, 
where $\mathcal{D}_\text{context}$ is a domain-informed external task distribution for interpolating to $\mathcal{D}_\text{train}$, and $\mathcal{D}_\text{mvalid}$ is a meta-validation set used to guide the interpolation function.
Inspired by \cite{metainterpolation}, we introduce a permutation invariant learnable set function~\citep{deepsets,set-transformer} \(\mu_\lambda\) as a mixer (interpolator), which learns to mix each point from $x_i \sim \mathcal{D}_\text{train}$  with the context points $\{c_{ij}\}_{j=1}^{m_i}$ in a way that densifies $\mathcal{D}_\text{train}$, where
\[
(x_i, y_i) \sim \mathcal{D}_\text{train}, \quad \{c_{ij}\}_{j=1}^{m_i} \sim \mathcal{D}_\text{context}, \quad i\in \{1, \dots, B\}\]
and $B$ denotes the minibatch size, $m_i \sim \mathcal{U}_{\text{int}}(0, M)$ where $M$ controls the maximum number of context samples drawn from $\mathcal{D}_\text{context}$ for each minibatch. Given a feature dimension $D$, for each $i$, the input consists of, $x_i \in \mathbb{R}^{B \times 1 \times D}$ and  $\{ c_{ij} \}_{j=1}^{m_i} \in \mathbb{R}^{B \times 1 \times D}$, where the set $\{ c_{ij} \}_{j=1}^{m_i}$ can be organized into a tensor $C_i \in \mathbb{R}^{B \times m_i \times D}$.

Overall, our model has two main components: (1) 
a meta-learner \( f_{\theta_l} \), which is a standard MLP at the \( l^{\text{th}} \) layer, that maps input data \( x_{i}^{(l-1)} \in \mathbb{R}^{B \times 1 \times D} \) to the feature space of the \( (l+1)^{\text{th}} \) layer, producing \( x_{i}^{(l)} = f_{\theta_l}(x_{i}^{(l-1)}) \), 
and (2) a learnable set function $\mu_\lambda$ which mixes $x_{i}^{{(l_\text{mix})}}$ and $C_{i}^{{(l_\text{mix})}}$ as a set and outputs a single pooled representation $\tilde{x}_i^{{(l_\text{mix})}} = \mu_{\lambda}(\{x_{i}^{{(l_\text{mix})}}, C_{i}^{{(l_\text{mix})}}\}) \in \mathbb{R}^{B \times 1 \times H}$, where $H$ is the hidden dimension and $l_\text{mix}$ is the layer where the mixing happens. The full model structure with $L$ layers can be expressed as \[
\hat{f}_{\theta, \lambda} := f_{\theta_L} \circ \cdots \circ f_{\theta_{l_\text{mix}+1}} \circ \mu_{\lambda} \circ f_{\theta_{l_\text{mix}-1}} \circ \cdots \circ f_{\theta_1}.
\]

\setlength{\tabcolsep}{12pt}
\begin{table*}[t]
\small
    \centering
    \caption{Merck Molecular Activity Challenge. We report performance (MSE $\downarrow$) on datasets with the large test-time covariate shift (HIVPROT, DPP4, NK1). All entries show the mean and standard errors calculated over 10 training runs. The best performing models are highlighted in bold. 'outlier exposure' indicates whether outlier exposure is enabled, and 'bilevel' denotes the use of bilevel optimization.}
    \label{tab:merck_results}
    \vspace{1pt}
    \begin{adjustbox}{max width=\textwidth}
    \begin{tabular}{l|cc|cc|cc}
    \toprule
    \small
    \multirow{2}{*}{Model} & \multicolumn{2}{c|}{HIVPROT} & \multicolumn{2}{c|}{\textsc{DPP4}} & \multicolumn{2}{c}{\textsc{NK1}}\\
    & count vector & bit vector & count vector & bit vector & count vector & bit vector \\
    \midrule
    $L_1$-Regression & $1.137\pms{.000}$ & $0.714\pms{.000}$ & $1.611\pms{.000}$ & $1.130\pms{.000}$ & $0.482\pms{.000}$ & $0.442\pms{.000}$ \\
    $L_2$-Regression & $0.999\pms{.000}$ & $0.723\pms{.000}$ & $1.495\pms{.000}$ & $1.143\pms{.000}$ & $0.498\pms{.000}$ & $0.436\pms{.000}$ \\
    Random Forest & $0.815\pms{.009}$ & $0.834\pms{.010}$ & $1.473\pms{.008}$ & $1.461\pms{.012}$ & $0.458\pms{.002}$ & $0.438\pms{.002}$ \\
    MLP & $0.768\pms{.014}$ & $2.118\pms{.015}$ & $1.393\pms{.024}$ & $1.094\pms{.029}$ & $0.443\pms{.007}$ & $0.399\pms{.006}$ \\

Mixup & $0.764\pms{0.008}$ & $0.691\pms{0.022}$ & $1.439\pms{0.021}$ & $1.212\pms{0.012}$ & $0.481\pms{0.002}$ & $0.479\pms{0.003}$ \\
Mixup (w/ outlier exposure) & $0.748\pms{0.01}$ & $0.677\pms{0.015}$ & $1.384\pms{0.012}$ & $1.224\pms{0.016}$ & $0.442\pms{0.005}$ & $0.443\pms{0.005}$ \\

Manifold Mixup & $0.88\pms{0.023}$ & $0.898\pms{0.022}$ & $1.414\pms{0.021}$ & $1.367\pms{0.04}$ & $0.432\pms{0.005}$ & $0.499\pms{0.013}$ \\
    
Manifold Mixup (w/ bilevel) & $0.484\pms{0.011}$ & $0.804\pms{0.086}$ & 
$1.19\pms{0.05}$ & $1.217\pms{0.068}$ & $0.43\pms{0.013}$ & $0.536\pms{0.039}$ \\
    
    Q-SAVI & $0.682\pms{.019}$ & $0.664\pms{.028}$ & $1.332\pms{.017}$ & $1.028\pms{.027}$ & $0.436\pms{.007}$ & $0.387\pms{.012}$ \\

    \midrule

    \textbf{Ours (Deepsets)} & 
$0.555 \pm 0.096$ & 
\cellcolor[HTML]{E6F2FF}$\mathbf{0.364 \pm 0.018}$ & 
\cellcolor[HTML]{E6F2FF}$\mathbf{0.984 \pm 0.018}$ & 
\cellcolor[HTML]{E6F2FF}$\mathbf{0.963 \pm 0.017}$ & 
$0.455 \pm 0.016$ & 
\cellcolor[HTML]{E6F2FF}$\mathbf{0.376 \pm 0.008}$ \\

\textbf{Ours (Set Trans.)} & \cellcolor[HTML]{E6F2FF}$\mathbf{0.39 \pm 0.011}$ & 
$0.726 \pm 0.159$ & 
$1.121 \pm 0.037$ & 
$0.986 \pm 0.021$ & 
\cellcolor[HTML]{E6F2FF}$\mathbf{0.429 \pm 0.01}$ & 
$0.397 \pm 0.015$ \\

    \bottomrule
    \end{tabular}
    \end{adjustbox}
\end{table*}

 We utilize bilevel optimization for training meta-learner $f_{\theta_l}$, and treat the set function parameter $\mu_\lambda$ as a hyperparameter to be optimized in the outer loop \citep{hypergrad}.
 As shown in \cref{tab:ours_ablation} (w/o bilevel optimization), simply optimizing the meta-learner parameters $\theta$ and the set function parameters $\lambda$ jointly can lead to overfitting to the task distribution and harms test-time generalization.
 Following the setting of \citep{hypergrad}, during training, we only update the parameter $\theta$ in the inner loop and  we only update the parameter $\lambda$ in the outer loop (see \cref{fig:model_structure_b} for the detailed model structure of the bilevel optimization).
 
In the inner loop, the model accepts $x_{i} \in \mathbb{R}^{B \times 1 \times H}$ and $C_i \in \mathbb{R}^{B \times m_i \times H}$ and the set encoder $\mu_\lambda$ mixes $\{x_{i}^{{(l_\text{mix})}}, C_{i}^{{(l_\text{mix})}}\}$ and outputs $\tilde{x}_i^{{(l_\text{mix})}} \in \mathbb{R}^{B \times 1 \times H}$. Since $C_i$ is used to introduce a domain-informed external context to densify $\mathcal{D}_\text{train}$, we utilize the original label $y_i$ from $\mathcal{D}_{\text{train}}$ to train the task learner parameters $f_{\theta_l}$, with the mixed $\tilde{x}_i^{{(l_\text{mix})}}$.

In the outer loop, we train the set encoder using hypergradient~\citep{hypergrad}, which aims to minimize a meta-validation loss \( L_V(\lambda, \theta^*(\lambda)) \), where the model parameters \( \theta^*(\lambda) \) are the solution to the inner training objective $\theta^*(\lambda) = \arg\min_\theta L_T(\theta, \lambda)$
and \( L_T(\theta, \lambda) \) denotes the \textit{training loss} computed on a labeled dataset \( \mathcal{D}_{\text{train}} \), potentially regularized or influenced by the hyperparameters \( \lambda \). This inner loss \( L_T \) governs the optimization of model parameters $\theta$, while \( L_V \) evaluates generalization performance on a meta-validation set \( \mathcal{D}_{\text{mvalid}} \), and guides the update of \( \lambda \).
In order for the hypergradient to train the set encoder $\mu_\lambda$ in a way that guides the overall model $\hat{f}_{\theta, \lambda}$ toward robustness under covariate shift, we construct $\mathcal{D}_{\text{mvalid}}$ as 
\[
\{x_{i,k}^{(\text{mvalid})}\}_{k=1}^{K} \sim \mathcal{D}_\text{unlabeled}, \;\; y_{i,k}^{( \text{mvalid})} \sim \mathcal{N}(0, 1),
\] 
where $K$ is a hyperparameter of the samples drawn from $\mathcal{D}_\text{unlabeled}$ for each minibatch.
In \cref{tab:ours_ablation}, we empirically show that using a random $y_{i,k}^{( \text{mvalid})} \sim \mathcal{N}(0, 1)$ achieves better or comparable performance to using a labeled $y_{i,k}^{( \text{mvalid})} \sim \mathcal{D}_{\text{oracle}}$ corresponding to the real label for $x_{i,k}^{(\text{mvalid})}$ from the unlabeled dataset which has been labeled by an oracle. 
This setting aligns with the role of \( L_V \) in guiding the set function \( \mu_{\lambda} \) to generalize to out-of-distribution data, as sampling pseudo-labels from \( \mathcal{N}(0, 1) \) introduces controlled label noise that regularizes the model.
The outer loop loss $L_V$ can then be expressed as a function of two variables $L_V(\lambda, \theta^*(\lambda))$, and therefore, the hypergradient is given by,
\[
\frac{d L_V}{d \lambda} = \frac{\partial L_V}{\partial \lambda} + \frac{\partial L_V}{\partial \theta} \cdot \frac{d \theta^*(\lambda)}{d \lambda}
\]
and Implicit Function Theorem (IFT) is applied to compute \( \frac{d \theta^*(\lambda)}{d \lambda} \) without differentiating through the entire optimization trajectory:
\[
\frac{d \theta^*(\lambda)}{d \lambda} = -\left( \frac{\partial^2 L_T}{\partial \theta^2} \right)^{-1} \cdot \frac{\partial^2 L_T}{\partial \theta \partial \lambda}.
\]

Luckily, the inverse Hessian vector product can be approximated and computed efficiently using automatic differentiation and Neuman series iterations~\citep{hypergrad}.
In \cref{analysis}, we display how the model $\hat{f}_{\theta, \lambda}$ effectively uses $\mathcal{D}_\text{context}$ to densify the data distribution around $\mathcal{D}_\text{train}$.

As the mixing of $\mathcal{D}_{\text{train}}$ and $\mathcal{D}_{\text{context}}$ is meant to densify the sparse training data, we only perform the mixing at train time. At test time, the set function $\mu_\lambda$ receives an input as a singleton set and performs no mixing with any exogeneous distribution (see \cref{fig:model_structure_a} for train and test-time diagram). 

\section{Experiments}
\label{experiments}

\setlength{\tabcolsep}{12pt}
\begin{table*}[t]
\small
    \centering
\caption{Ablation study of \textbf{Ours}. 'ctx' indicates the use of context points, while 'bilevel' denotes the application of bilevel optimization. For $y^{(\text{mvalid})}$, 'rand' refers to sampling from a standard normal distribution, whereas 'real' refers to sampling from the oracle distribution.}

    \label{tab:ours_ablation}
    \vspace{1pt}
    \begin{adjustbox}{max width=\textwidth}
    \begin{tabular}{lccc|cc|cc|cc}
    \toprule
    \small
    \multirow{2}{*}{Model} & \multirow{2}{*}{w/ ctx.} & \multirow{2}{*}{w/ bilevel} & \multirow{2}{*}{$y^{(\text{mvalid})}$} & \multicolumn{2}{c|}{HIVPROT} & \multicolumn{2}{c|}{\textsc{DPP4}} & \multicolumn{2}{c}{\textsc{NK1}}\\
    & & & & count vector & bit vector & count vector & bit vector & count vector & bit vector \\
    \midrule
    \textbf{Ours (Deepsets)} & \cmark & \xmark & rand & $1.147 \pm 0.055$ & 
$1.012 \pm 0.025$ & 
$1.342 \pm 0.019$ & 
$1.344 \pm 0.039$ & 
$0.45 \pm 0.01$ & 
$0.494 \pm 0.006$ \\

    \textbf{Ours (Set Trans.)} &\cmark & \xmark & rand & $1.193 \pm 0.038$ & 
$1.008 \pm 0.03$ & 
$1.34 \pm 0.026$ & 
$1.359 \pm0.013$ & 
$0.482 \pm 0.002$ & 
$0.487 \pm 0.009$ \\

    \textbf{Ours (MLP)} & \xmark & \xmark & rand & $1.043 \pm 0.013$ & 
$1.049 \pm 0.013$ & 
$1.42 \pm 0.009$ & 
$1.269 \pm 0.017$ & 
$0.435 \pm 0.002$ & 
$0.466 \pm 0.004$ \\
       \textbf{Ours (Deepsets)} & \cmark & \cmark & real &  
$0.506 \pm 0.061$ & 
$0.381 \pm 0.018$ & 
\cellcolor[HTML]{E6F2FF}$\mathbf{0.970 \pm 0.012}$ & 
$0.964 \pm 0.024$ & 
$0.495 \pm 0.037$ & 
$0.395 \pm 0.014$ \\
    \textbf{Ours (Set Trans.)} & \cmark & \cmark & real & $0.401 \pm 0.02$ & 
$0.811 \pm 0.111$ & 
$1.123 \pm 0.024$ & 
$0.975 \pm 0.016$ & 
$0.449 \pm 0.021$ & 
$0.401 \pm 0.017$ \\
\midrule

    \textbf{Ours (Deepsets)}  & \cmark & \cmark & rand & 
$0.555 \pm 0.096$ & 
\cellcolor[HTML]{E6F2FF}$\mathbf{0.364 \pm 0.018}$ & 
$0.984 \pm 0.018$ & 
\cellcolor[HTML]{E6F2FF}$\mathbf{0.963 \pm 0.017}$ & 
$0.455 \pm 0.016$ & 
\cellcolor[HTML]{E6F2FF}$\mathbf{0.376 \pm 0.008}$ \\

\textbf{Ours (Set Trans.)} & \cmark & \cmark & rand & \cellcolor[HTML]{E6F2FF}$\mathbf{0.39 \pm 0.011}$ & 
$0.726 \pm 0.159$ & 
$1.121 \pm 0.037$ & 
$0.986 \pm 0.021$ & 
\cellcolor[HTML]{E6F2FF}$\mathbf{0.429 \pm 0.01}$ & 
$0.397 \pm 0.015$ \\




    \bottomrule
    \end{tabular}
    \end{adjustbox}
\end{table*}

We evaluate on the Merck Molecular Activity Challenge (Merck) dataset \citep{merck_activity}, a benchmark for molecular property prediction. The dataset contains bioactivity measurements for therapeutic targets, with molecules represented by high-dimensional chemical descriptors. It comprises 15 distinct datasets with a regression target predicting a molecular activity measurement in (possibly) different units, enabling evaluation across multiple distribution shifts. The Merck dataset reflects real-world drug discovery scenarios where promising candidates lie outside known chemical spaces, making generalization challenging. Molecular inputs are encoded either as \textit{bit vectors}, indicating the presence or absence of substructures, or \textit{count vectors}, which reflect the frequency of substructures—providing varying granularity of chemical information.

Following the prior work \citep{qsavi}, for $\mathcal{D}_\text{train}$, we focus on the three subsets (HIVPORT, DPP4, NK1) that exhibit the greatest distributional shift from and treat the remaining 12 datasets as $\mathcal{D}_\text{unlabeled}$ and exclude their labels from the training process. For fair comparison with baselines, we randomly select some portion ($K$ for each minibatch) of $\mathcal{D}_\text{unlabeled}$ as $\mathcal{D}_\text{mvalid}$ and sample $\mathcal{D}_\text{context}$ from $\mathcal{D}_\text{unlabeled}$ ($m_i$ for each minibatch). As $\mathcal{D}_{\text{mvalid}}$ is randomly labeled, we ensure that all baselines see the same amount of labeled data. 


In addition to the baselines presented in \citep{qsavi}, we further implement and evaluate two related state-of-the-art interpolation methods—Mixup \citep{mixup} and Manifold Mixup \citep{manifold_mixup}—to empirically demonstrate the capability of our set function $\mu_{\lambda}$. As mixup performs the mixing procedure in the input space, there are no parameters to train for bilevel optimization. We therefore report two variants of mixup (plain and outlier exposure) where outlier exposure is exposed to the same $\mathcal{D}_{\text{mvalid}}$ with random labels as described above. Manifold mixup performs the mixup procedure in the feature space between layers, therefore we can straightforwardly apply our bilevel optimization procedure to manifold mixup. Therefore, we report two variants of manifold mixup (plain and bilevel optimization).

\begin{figure*}[t]
\vspace{-1em}
\centering
\begin{subfigure}[b]{0.32\linewidth}
    \centering
    \includegraphics[width=\linewidth]{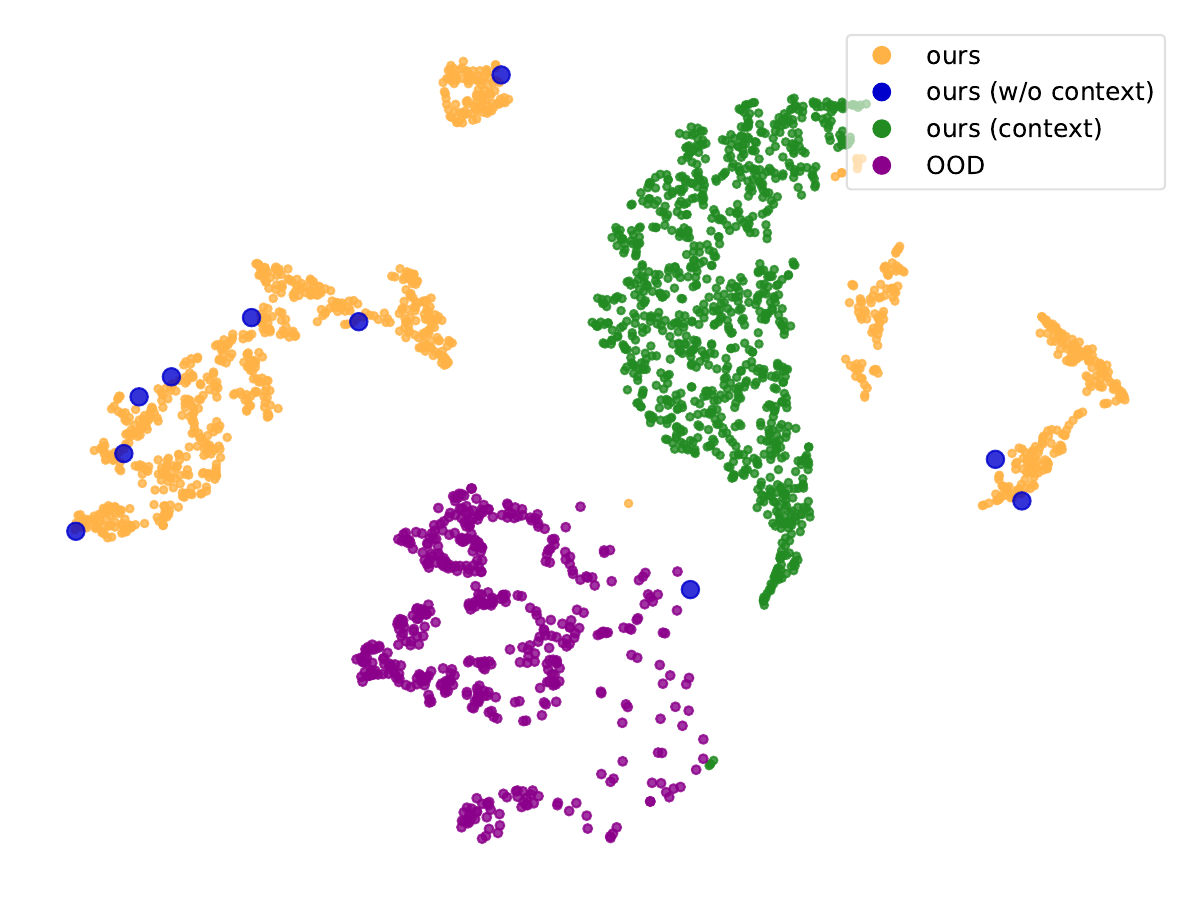}
    \caption{\textbf{Ours}}
    \label{fig:tsne_ours}
\end{subfigure}
\hfill
\begin{subfigure}[b]{0.32\linewidth}
    \centering
    \includegraphics[width=\linewidth]{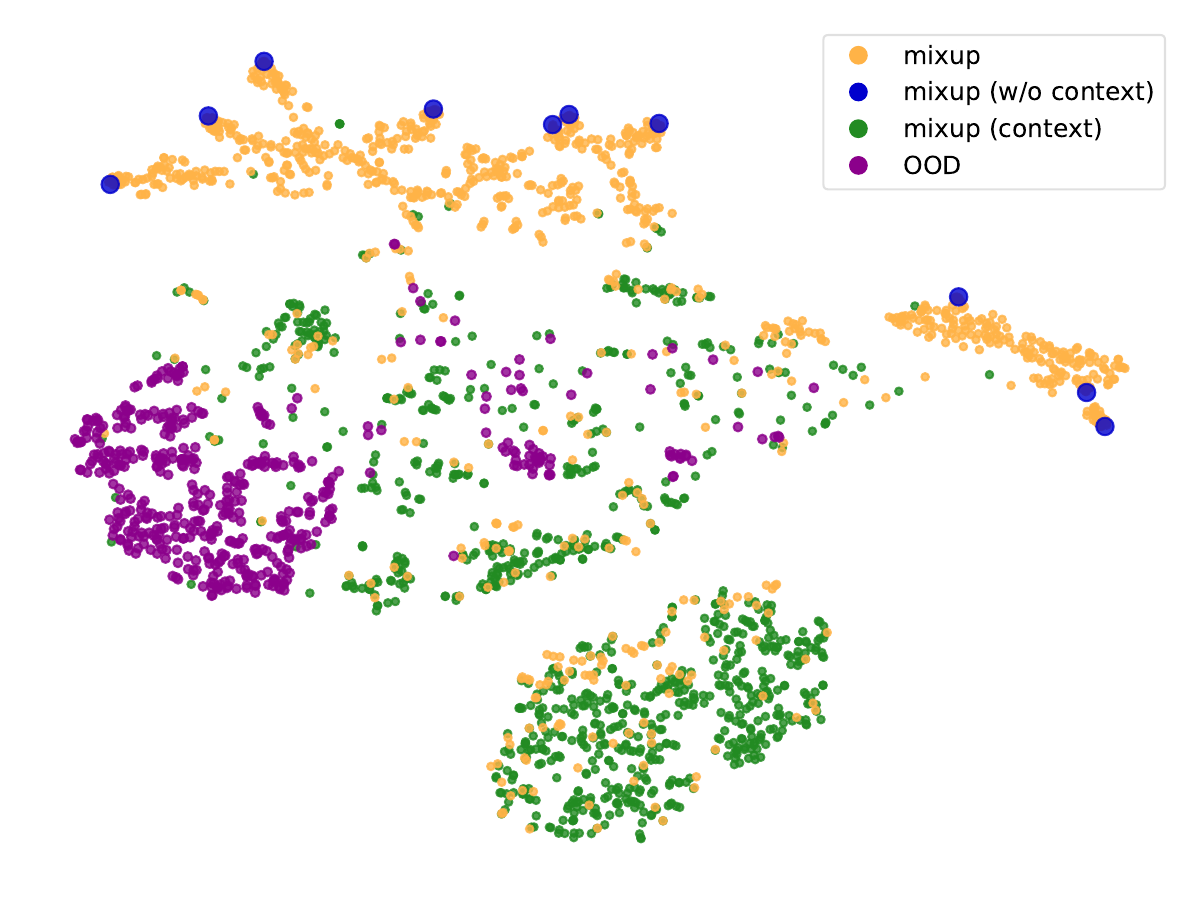}
    \caption{\textbf{Mixup (w/ outlier exposure)}}
    \label{fig:tsne_mixup}
\end{subfigure}
\hfill
\begin{subfigure}[b]{0.32\linewidth}
    \centering
    \includegraphics[width=\linewidth]{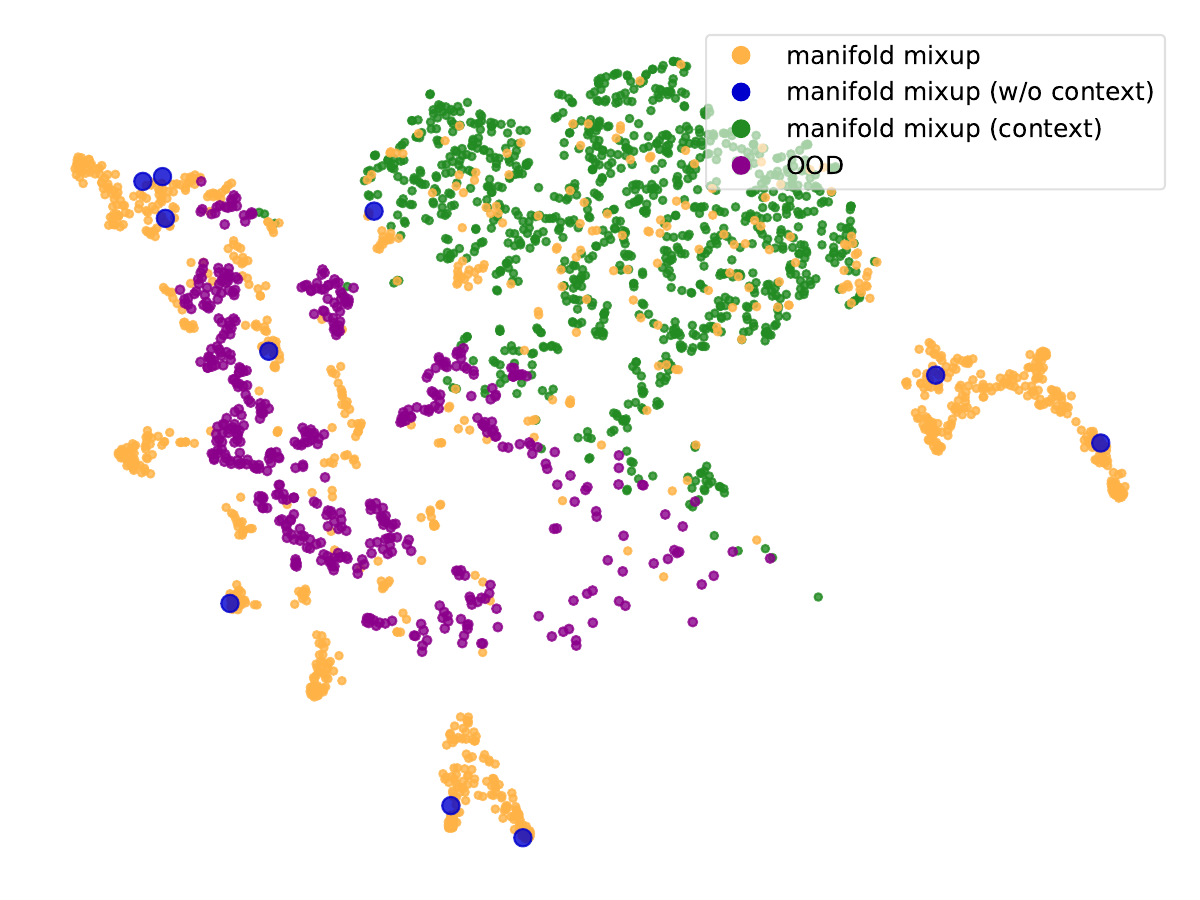}
    \caption{\textbf{Manifold Mixup (w/ bilevel optim.)}}
    \label{fig:tsne_manmixup}
\end{subfigure}
\vspace{-0.5em}
\caption{\small \textbf{t-SNE visualization of DPP4 (bit) dataset from the penultimate layer across different methods.} All models were trained on $\mathcal{D}_\text{train}$, $\mathcal{D}_\text{context}$, and $\mathcal{D}_\text{mvalid}$. At test time, we evaluate each model on four input variants (orange, blue, green, purple) to analyze how the model utilizes $\mathcal{D}_\text{context}$ to achieve robustness under covariate shift and how it behaves on out-of-distribution (OOD) data.}

\label{fig:tsne_nk1_bit}
\end{figure*}

For Mixup, we set $l_\text{mix}=1$ so we mix $\{x_{i}^{{(l_\text{1})}}, C_{i}^{{(l_\text{1})}}\}$
and for Manifold Mixup, we set $l_\text{mix} > 1$, so we mix $\{x_{i}^{{(l_\text{mix})}}, C_{i}^{{(l_\text{mix})}}\}$ after passing at least one layer $f_{\theta_l}$ (For further implementation details, please see~\cref{app:experiment-details_tab1}). 
When evaluating with bilevel optimization, we follow our setting and update $\theta$ and $\lambda$ separately in each loop using hypergrad \citep{hypergrad}, and for the setting without bilevel optimization, we jointly update $\theta$ and $\lambda$ using a single optimizer.
As shown in \cref{tab:merck_results}, when either leveraging Deepsets \citep{deepsets} or Set Transformer \citep{set-transformer} as the mixer \(\mu_\lambda\), our method  significantly improves MSE compared to the baselines, demonstrating greater robustness of our model toward covariate shift.

Furthermore, we present a series of ablation studies in \cref{tab:ours_ablation} to analyze the contributions of each component of our method (Implementation details can be found in \cref{app:experiment-details_tab2}).
Specifically, to validate the effectiveness of \( \mathcal{D}_\text{context} \) and our bilevel optimization~\citep{hypergrad} under \( y_{i,k}^{(\text{mvalid})} \sim \mathcal{N}(0, 1) \), we compare our approach against the following variants: (1) using $\mathcal{D}_\text{context}$ without bilevel optimization, (2) excluding both $\mathcal{D}_\text{context}$ and bilevel optimization, and (3) using the real label for $y_{i,k}^{( \text{mvalid})}$. 
The results indicate that, in order to achieve robust generalization across all datasets, it is crucial to leverage bilevel optimization with $\mathcal{D}_\text{context}$. Especially, the dataset HIVPROT, which has the highest degree of label and covariate shift compared to others \citep{qsavi} (Table 3), showed a detrimental performance drop ($62.6\%\text{–}67.3\%$ for the count vector, and $63.8\%\text{–}65.3\%$ for the bit vector) in settings (1,2). Moreover, using $y_{i,k}^{( \text{mvalid})}\sim \mathcal{N}(0, 1)$ in $L_V$ for training the set function $\mu_{\lambda}$ resulted in comparable or better performance than using the true label associated with $x_{i,k}^{( \text{mvalid})}$, validating our choice of $L_V$ and its contribution to improved robustness under covariate shift.

\vspace{-1em}
\section{Analysis}
\label{analysis}

%
To visually assess how our model leverages $\mathcal{D}_\text{context}$ to densify $\mathcal{D}_\text{train}$, and to compare model behaviors on out-of-distribution (OOD) inputs, we visualize the penultimate-layer embeddings $\mathbf{z}_i = x_i^{(L-1)}$
, using t-SNE~\citep{tsne} across three models that interpolate $\mathcal{D}_\text{train}$ with $\mathcal{D}_\text{context}$: Ours, Mixup (w/ outlier exposure), and Manifold Mixup (w/ bilevel optim.), all trained on the DPP4 bit-vector dataset.
We randomly sample $x_i, y_i \sim \mathcal{D}_\text{train}$, $C_i=\{c_{ij}\}_{j=1}^{100} \sim \mathcal{D}_\text{context}$, $x_i^{\text{ood}}
 \sim \mathcal{D}_\text{ood}$, where $\mathcal{D}_\text{ood}$ signifies an out-of-distribution dataset which was never seen during training.
In \cref{fig:tsne_nk1_bit}, each color indicates a penultimate-layer embedding $\mathbf{z} \in \mathbf{Z}$ under different input configurations.
 We denote the embeddings as follows: $\mathbf{Z}_{\text{joint}}$ for \textbf{\textcolor{orangehighlight}{orange}} ($\{x_i, C_i\}$), $\mathbf{Z}_{\text{input}}$ for \textbf{\textcolor{deepblue}{blue}} ($x_i$ alone), $\mathbf{Z}_{\text{context}}$ for \textbf{\textcolor{forestgreen}{green}} ($C_i$ alone), and $\mathbf{Z}_{\text{ood}}$ for \textbf{\textcolor{darkmagenta}{purple}} (OOD samples $x_i^{\text{ood}}$). Additional t-SNE results and details are in \cref{tSNE_detail}.
 
\textbf{Effect of interpolating $\mathcal{D}_\text{train}$ with $\mathcal{D}_\text{context}$.}
In \cref{fig:tsne_ours}, we can see that our model successfully uses $C_i$ to densify each train input $x_i$, based on the fact that $\mathbf{Z}_{\text{joint}}$ is concentrated in the vicinity of $\mathbf{Z}_{\text{input}}$ and is located in a completely separate cluster from both $\mathbf{Z}_{\text{context}}$ and $\mathbf{Z}_{\text{ood}}$. On the other hand, as shown in \cref{fig:tsne_mixup,fig:tsne_manmixup}, both Mixup (w/ outlier exposure) and Mandifold Mixup (w/ bilevel optim.) show extensive overlap between $\mathbf{Z}_{\text{joint}}$, $\mathbf{Z}_{\text{context}}$, and $\mathbf{Z}_{\text{ood}}$.

Unlike Mixup (w/ outlier exposure) and Manifold Mixup (w/ bilevel optimization), our method shows much clearer separation between data distributions. This shows that our model effectively distinguishes seen from unseen distributions while learning structured, robust latent representations that enhance generalization under covariate shift, consistent with the results in \cref{tab:merck_results}.

\vspace{-1em}
\section{Conclusion}
\label{conclusion}
To overcome the challenge of test-time covariate shift, we propose a novel bilevel optimization method that densifies the training distribution using domain-informed unlabeled datasets via a learnable set function. During training, we utilize bilevel optimization which leverages noisy unlabeled data to guide our model towards robustness under covariate shift.
We validate our method on challenging real-world molecular property prediction with large covariate shifts, and visually demonstrate the effectiveness of interpolation method which shows well defined separation between data distributions. 
Our code is available at \url{https://github.com/JinA0218/drugood-densify}.

\bibliography{example_paper}
\bibliographystyle{icml2025}

\newpage
\appendix
\onecolumn

\section{Model Structure Overview}
\label{app:detailed_model_structure}
\begin{figure}[H]
\centering
\begin{subfigure}[b]{0.47\linewidth}
    \centering
    \includegraphics[width=\linewidth]{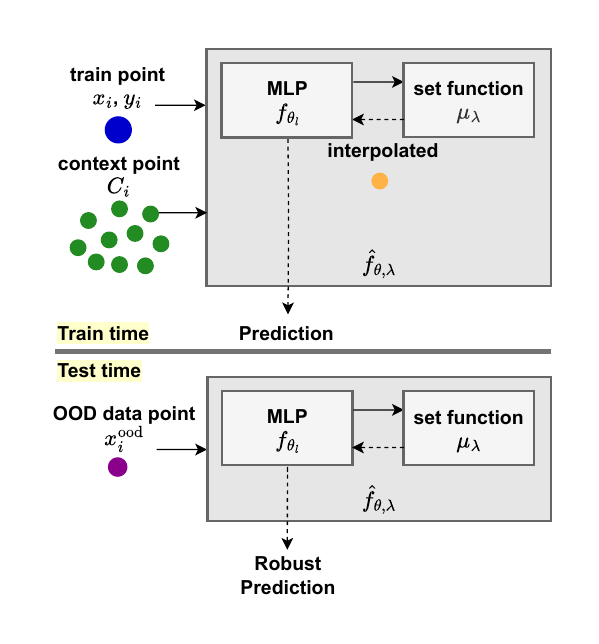}
    \caption{\textbf{Training and Test-time Framework}}
    \label{fig:model_structure_a}
\end{subfigure}
\hfill
\begin{subfigure}[b]{0.47\linewidth}
    \centering
    \includegraphics[width=\linewidth]{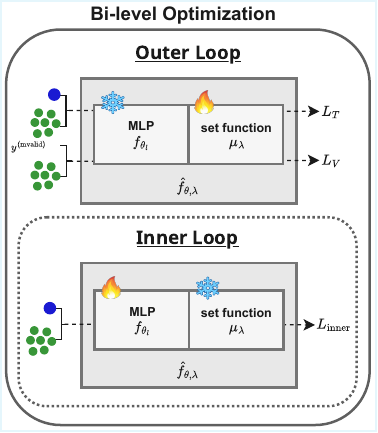}
    \caption{\textbf{Bilevel Optim. during Training}}
    \label{fig:model_structure_b}
\end{subfigure}
\vspace{-0.5em}
\caption{\small \textbf{Overview of our proposed model.} (a) During training, the model interpolates between a labeled train point $(x_i, y_i)$ and context point $C_i$ to learn robust representations. At test time, the model predicts on an OOD input using the learned meta learner $f_\theta$ and set function $\mu_\lambda$. (b) The model is trained via bilevel optimization, where the inner loop updates $\theta$ using the inner loss $L_{\text{inner}}$, while the outer loop updates $\lambda$ using the hypergradient computed from $L_T$ and $L_V$.
    }
\end{figure}

\section{Experimental Details}
All code and experiments are publicly available at \url{https://github.com/JinA0218/drugood-densify}.
\label{app:experiment-details}
\subsection{Implementation details of the baselines (\cref{tab:merck_results})}
\label{app:experiment-details_tab1}
In \cref{tab:merck_results}, we report the performance of $L_1$-Regression, $L_2$-Regression, Random Forest, MLP, and Q-SAVI by referring to the results from \citep{qsavi}. For Mixup (plain and with outlier exposure) and Manifold Mixup (plain and with bilevel optimization), we replace only our set function \( \mu_{\lambda} \) with the linear interpolation \( \mu_\alpha^{\text{lin}} \) and reduction function, defined as 
\[
\tilde{x}_i^{{(l_\text{mix})}} = f_{\text{reduce}}\left( \alpha x_i + (1 - \alpha) C_i \right), \quad \alpha \sim \mathcal{U}(0, 1), \ \alpha \in \mathbb{R}^{B \times 1 \times 1}, \ f_{\text{reduce}} \in \{\text{mean}, \text{sum}, \text{max}\},
\]
while keeping all other components unchanged. We apply the function \( f_{\text{reduce}} \) since \( C_i \in \mathbb{R}^{B \times m_i \times 1} \), and therefore a final reduction must be applied in order to reduce the set dimension $m_i$ to 1. We us \( \max \) for $f_{\text{reduce}}$ in \cref{tab:merck_results} as a sensible default. 

\textbf{Mixup and Manifold Mixup.} For the plain Mixup and Manifold Mixup (without bilevel or outlier exposure), we perform the mixing only utilizing $(x_i, y_i) \sim \mathcal{D}_\text{train}$ and $\{c_{ij}\}_{j=1}^{m_i} \sim \mathcal{D}_\text{context}$ without using ${D}_\text{mvalid}$, without any outlier exposure. Since this impelementation doesn't include bilevel optimization, we train each Mixup and Manifold Mixup with a single optimizer.

\textbf{Mixup (w/ outlier exposure).} To compare Mixup with outlier exposure, we implemented a variant of Mixup which is trained with $(x_i, y_i) \sim \mathcal{D}_\text{train}$, $\{c_{ij}\}_{j=1}^{m_i} \sim \mathcal{D}_\text{context}$ and $\{x_{i,k}^{(\text{mvalid})}\}_{k=1}^{K} \sim \mathcal{D}_\text{unlabeled}, \;\; y_{i,k}^{( \text{mvalid})} \sim \mathcal{N}(0, 1)$. As explained in \cref{experiments}, since it peforms the mixing proceduce in the input space there are no parameters to train, it is impossible to apply bilevel optimization. Therefore, when computing the loss, we do not follow the operation held in hypergradient~\citep{hypergrad} but instead add train loss and $L_V$, and update the model parameter with a single optimizer.

\textbf{Manifold Mixup (w/ bilevel optim.).} As described in \cref{experiments}, unlike Mixup, in Manifold Mixup, linear layers are included in $\mu_\alpha^\text{lin}$, so it is possible to straightforwardly apply bilevel optimization. Therefore, we follow the same procedure described in \cref{method}, and train using \( \mathcal{D}_\text{context} \) and \( \{x_{i,k}^{(\text{mvalid})}\}_{k=1}^{K} \sim \mathcal{D}_\text{unlabeled} \), with pseudo-labels \( y_{i,k}^{(\text{mvalid})} \sim \mathcal{N}(0, 1) \). Like our method, we also employ a separate optimizer for training \( \mu_\alpha^\text{lin} \).




\subsection{Implementation details of the ablation study (\cref{tab:ours_ablation})}
\label{app:experiment-details_tab2}
In \cref{tab:ours_ablation}, we perform three types ablation study of ours : 

\begin{enumerate}
    \item Using $\mathcal{D}_\text{context}$ without bilevel optimization.
    \item Excluding both $\mathcal{D}_\text{context}$ and bilevel optimization.
    \item Using the real label for $y_{i,k}^{( \text{mvalid})}$.
\end{enumerate}

Since ablation studies (1) and (3) include the set function \( \mu_{\lambda} \), we report the performance of each using both DeepSets~\citep{deepsets} and Set Transformer~\citep{set-transformer} as implementations of \( \mu_{\lambda} \).

\textbf{For ablation study (1)}, we train the model with $(x_i, y_i) \sim \mathcal{D}_\text{train}$, $\{c_{ij}\}_{j=1}^{m_i} \sim \mathcal{D}_\text{context}$ and $\{x_{i,k}^{(\text{mvalid})}\}_{k=1}^{K} \sim \mathcal{D}_\text{unlabeled}, \;\; y_{i,k}^{( \text{mvalid})} \sim \mathcal{N}(0, 1)$, and sum the inner loop train loss $L_T$ and outer loop loss $L_V$ to get the final loss for each iteration. Since there is no bilevel optimization in this setting, we update the model parameter of $f_\theta$ and $\mu_{\lambda}$ together using a single optimizer.

\textbf{For ablation study (2)},  we train the model with $(x_i, y_i) \sim \mathcal{D}_\text{train}$ and $\{x_{i,k}^{(\text{mvalid})}\}_{k=1}^{K} \sim \mathcal{D}_\text{unlabeled}, \;\; y_{i,k}^{( \text{mvalid})} \sim \mathcal{N}(0, 1)$, and add the inner loop train loss $L_T$ and the outer loop $L_V$ to get the final loss for each iteration. Since there is no context points, there is no interpolation with external distributions in this setting. Additionally, we update the model parameter of $f_\theta$ and $\mu_{\lambda}$ together using a single optimizer, so the effect of the outer loop loss $L_V$ does not result in any bilevel optimization. 

\textbf{For ablation study (3)}, we replace  $y_{i,k}^{( \text{mvalid})} \sim \mathcal{N}(0, 1)$ to a labeled $y_{i,k}^{( \text{mvalid})} \sim \mathcal{D}_{\text{oracle}}$ and perform the same process as our standard setting described in \cref{method}.

\subsection{Hyperparameter tuning details}
For hyperparameter settings, all methods (including ours, Mixup, Manifold Mixup, Mixup with outlier exposure, and Manifold Mixup with bilevel optimization) reported in \cref{tab:merck_results} used a learning rate of \( 1 \times 10^{-3} \) for \( f_\theta \), and a learning rate of \( 1 \times 10^{-5} \) for the mixers \( \mu_{\lambda} \) and \( \mu_\alpha^{\text{lin}} \), where the learning rate of \( \mu_\alpha^{\text{lin}} \) is only used for Manifold Mixup (w/ bilevel optim.).
The number of layers \( (L) \) was set to 3, the hidden dimension \( (H) \) to 64, the dropout rate to 0.5, and the optimizer to \texttt{adamwschedulefree}~\citep{adamsched}. For methods utilizing bilevel optimization, we set the number of inner-loop iterations to 10 and outer-loop iterations to 50. 
We furthure perform hyperparameter search for all the models with the same hyperparameter search space as shown in \cref{tab:qsavi_hyperparams}. Since Mixup, Manifold Mixup do not utilize $\mathcal{D}_\text{mvalid}$, the hyperparameter search was conducted only over $M$.

For \cref{tab:ours_ablation}, all ablation studies used the same hyperparameter search space as in \cref{tab:merck_results}.The hyperparameter search space is presented in \cref{tab:qsavi_hyperparams}.

\begin{table}[ht]
\centering
\caption{Hyperparameter search space.}
\begin{tabular}{ll}
\toprule
\textbf{Hyperparameter} & \textbf{Search Space} \\
\midrule
maximum number of context per batch ($M$) & 1, 4, 8 \\
number of mvalid per batch ($K$) & 1, 6, 8 \\
\bottomrule
\end{tabular}
\label{tab:qsavi_hyperparams}
\end{table}

\subsection{Computing resources}
All experiments are conducted on GPUs including the GeForce RTX 2080 Ti, RTX 3090, and RTX 4090.
\section{t-SNE Visualization Details}
\label{tSNE_detail}
\subsection{Embedding generation procedure}
\begin{figure}[t]
\centering
\begin{subfigure}[b]{0.32\linewidth}
    \centering
    \includegraphics[width=\linewidth]{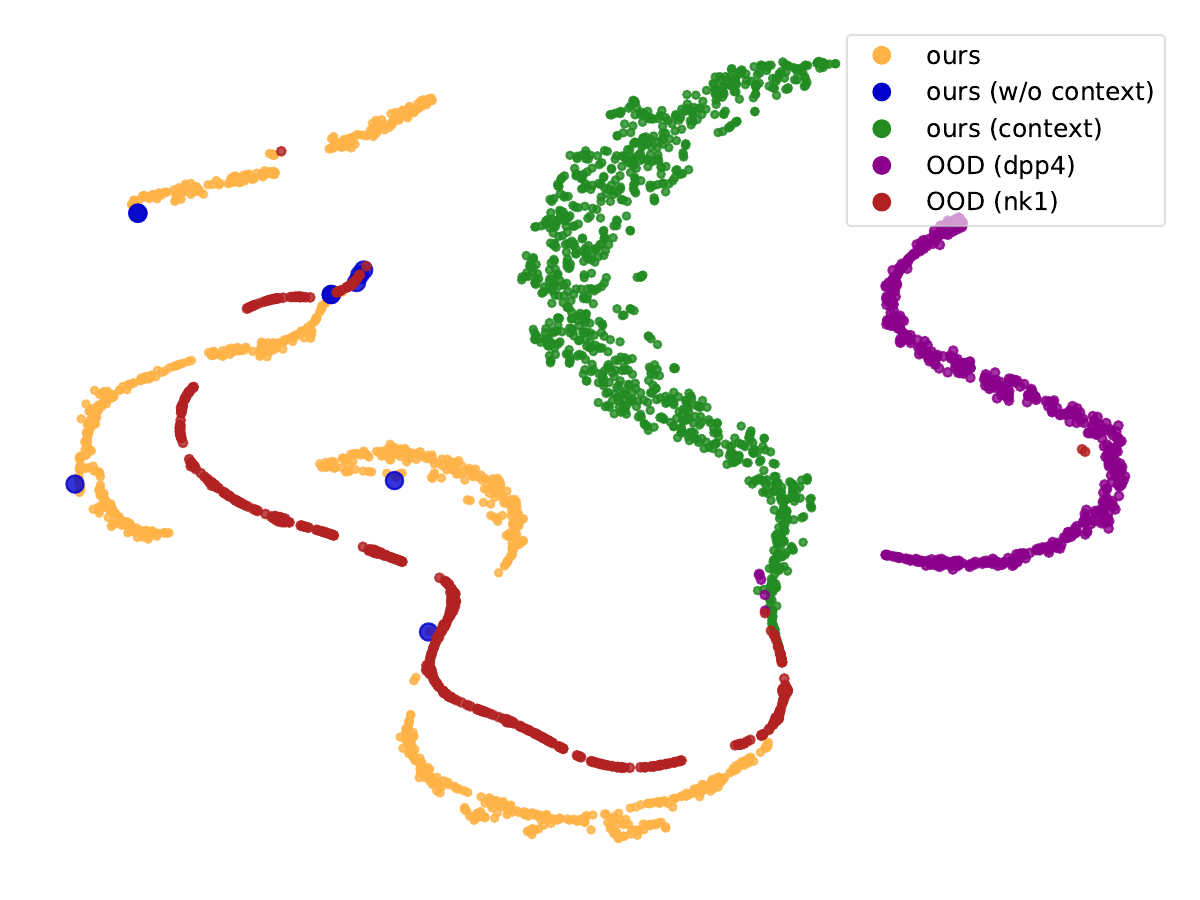}
    \caption{\textbf{Ours}}
\end{subfigure}
\hfill
\begin{subfigure}[b]{0.32\linewidth}
    \centering
    \includegraphics[width=\linewidth]{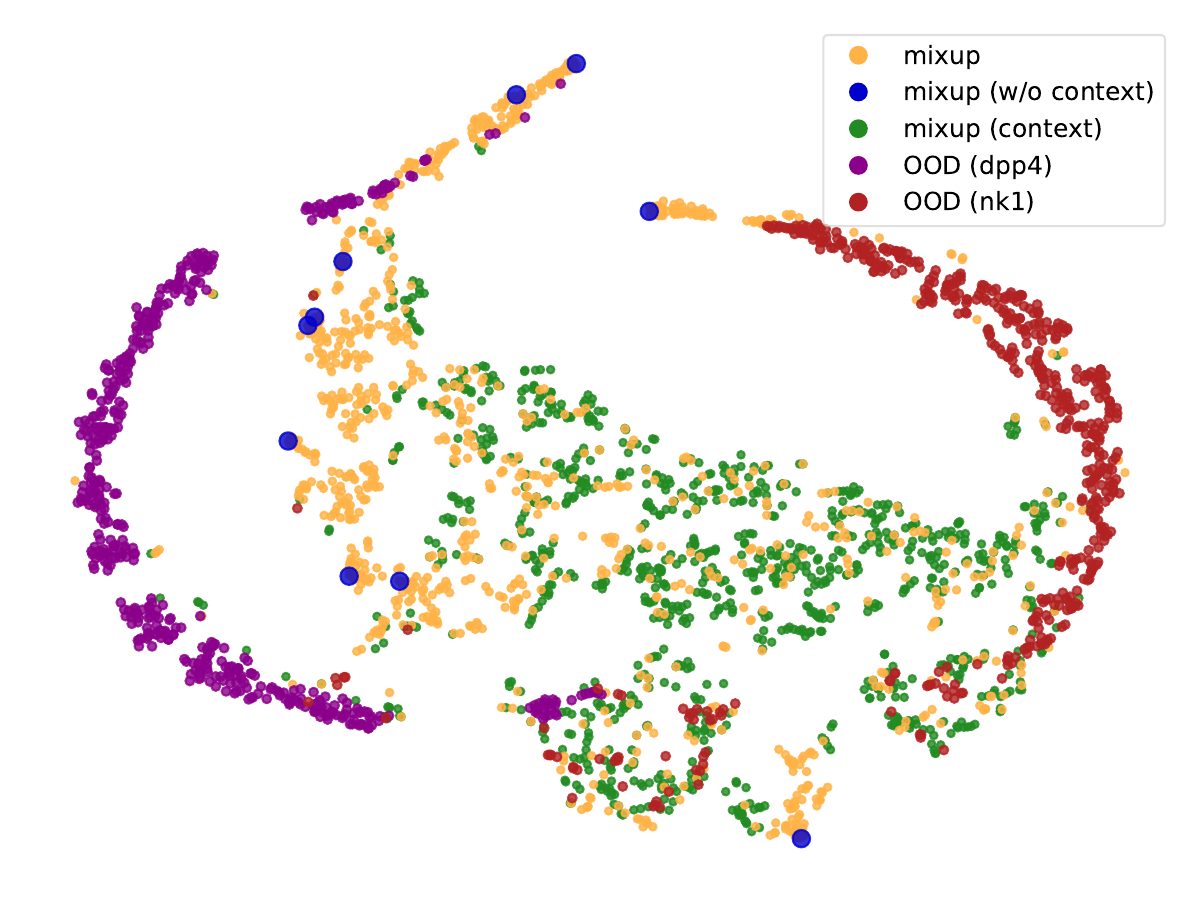}
    \caption{\textbf{Mixup (w/ outlier exposure)}}
\end{subfigure}
\hfill
\begin{subfigure}[b]{0.32\linewidth}
    \centering
    \includegraphics[width=\linewidth]{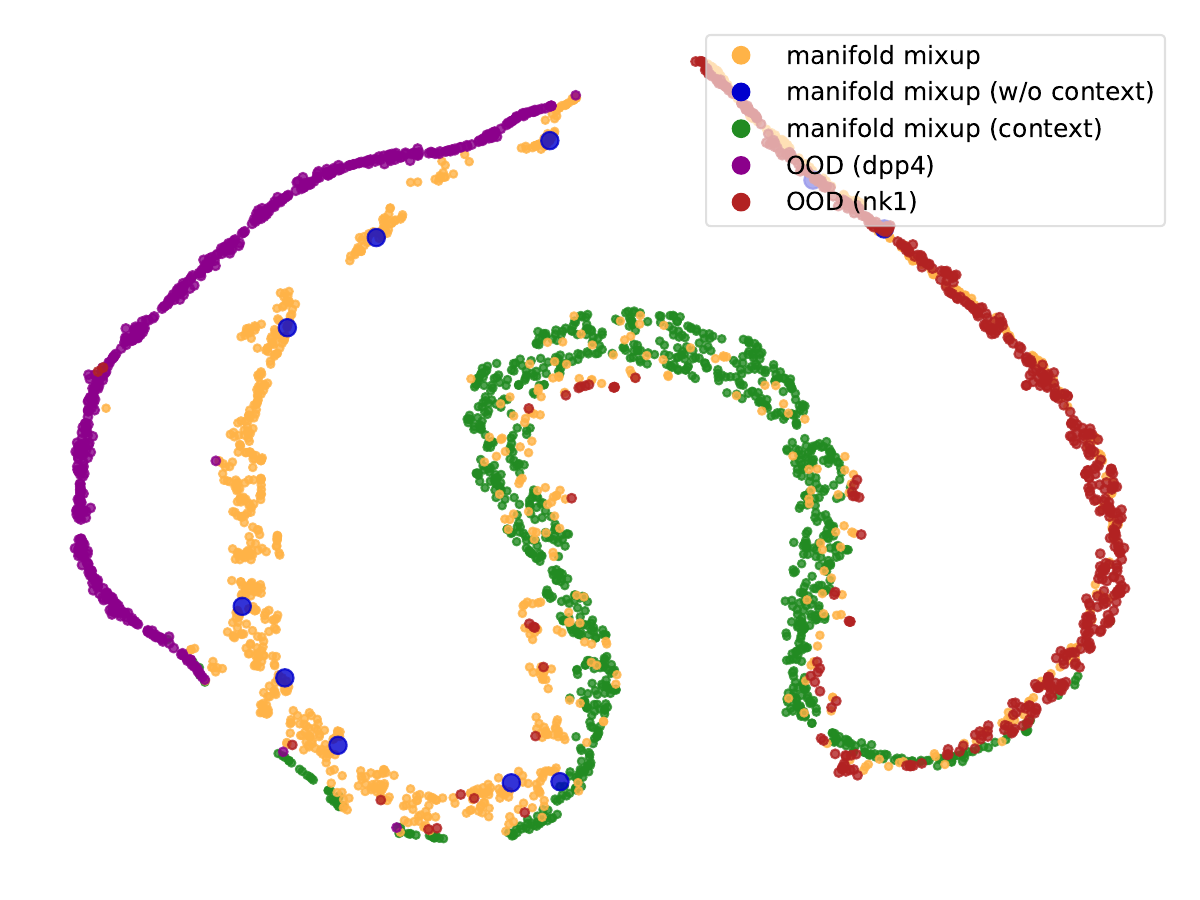}
    \caption{\textbf{Manifold Mixup (w/ bilevel optim.)}}
\end{subfigure}
\vspace{-0.5em}
\caption{\small \textbf{t-SNE visualization of the model trained on the HIVPROT (count) dataset}}
\label{fig:app_tsne_hivprot_count}
\end{figure}

\begin{figure}[t]
\centering
\begin{subfigure}[b]{0.32\linewidth}
    \centering
    \includegraphics[width=\linewidth]{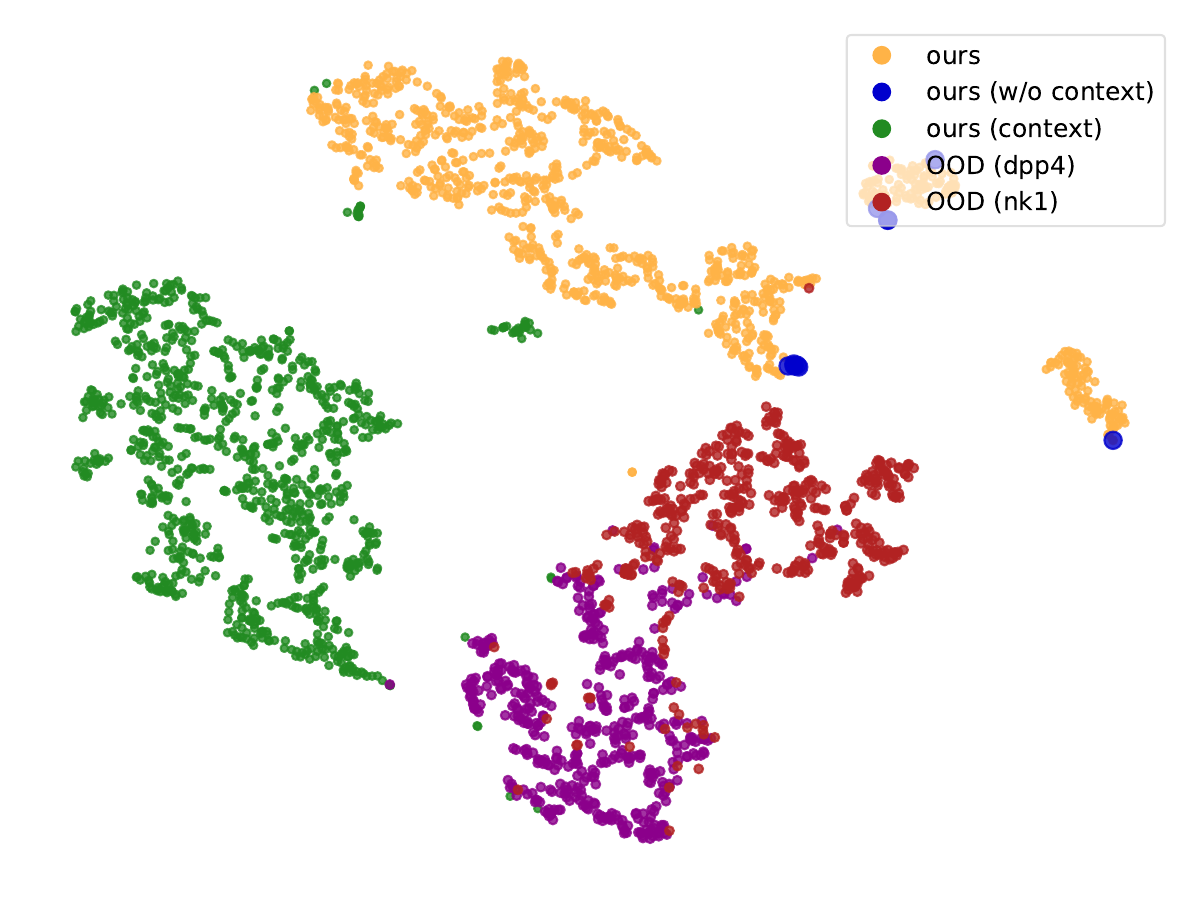}
    \caption{\textbf{Ours}}
\end{subfigure}
\hfill
\begin{subfigure}[b]{0.32\linewidth}
    \centering
    \includegraphics[width=\linewidth]{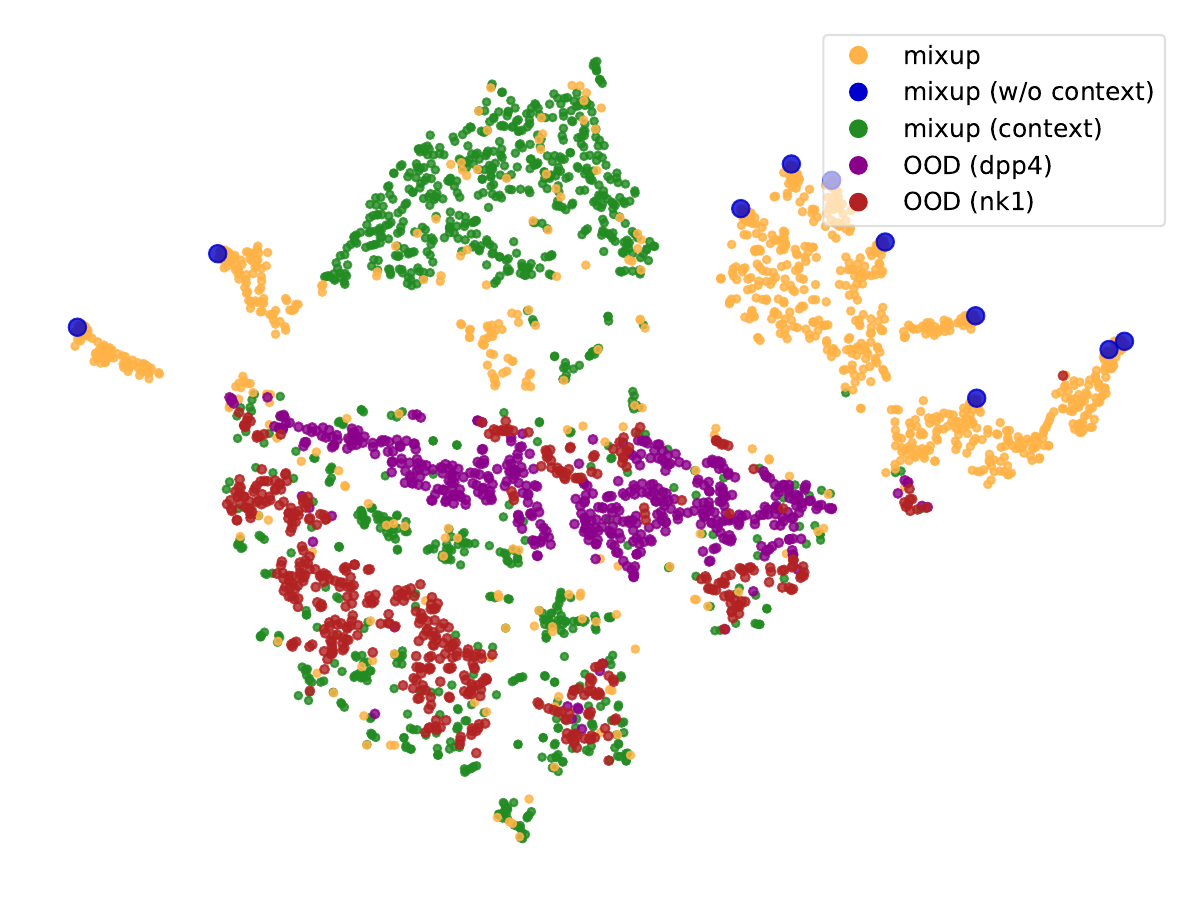}
    \caption{\textbf{Mixup (w/ outlier exposure)}}
\end{subfigure}
\hfill
\begin{subfigure}[b]{0.32\linewidth}
    \centering
    \includegraphics[width=\linewidth]{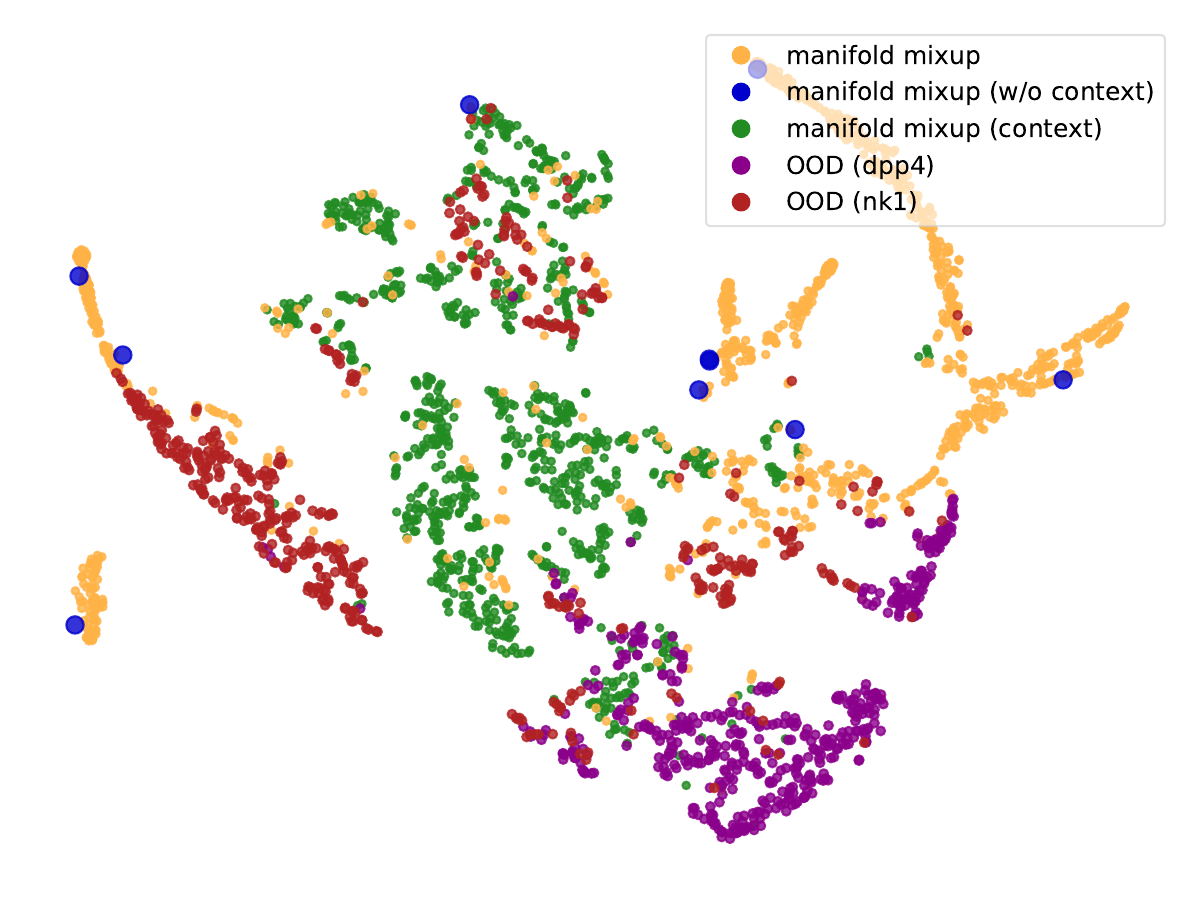}
    \caption{\textbf{Manifold Mixup (w/ bilevel optim.)}}
\end{subfigure}
\vspace{-0.5em}
\caption{\small \textbf{t-SNE visualization of the model trained on the HIVPROT (bit) dataset}}
\label{fig:app_tsne_hivprot_bit}
\end{figure}

\begin{figure}[t]
\centering
\begin{subfigure}[b]{0.32\linewidth}
    \centering
    \includegraphics[width=\linewidth]{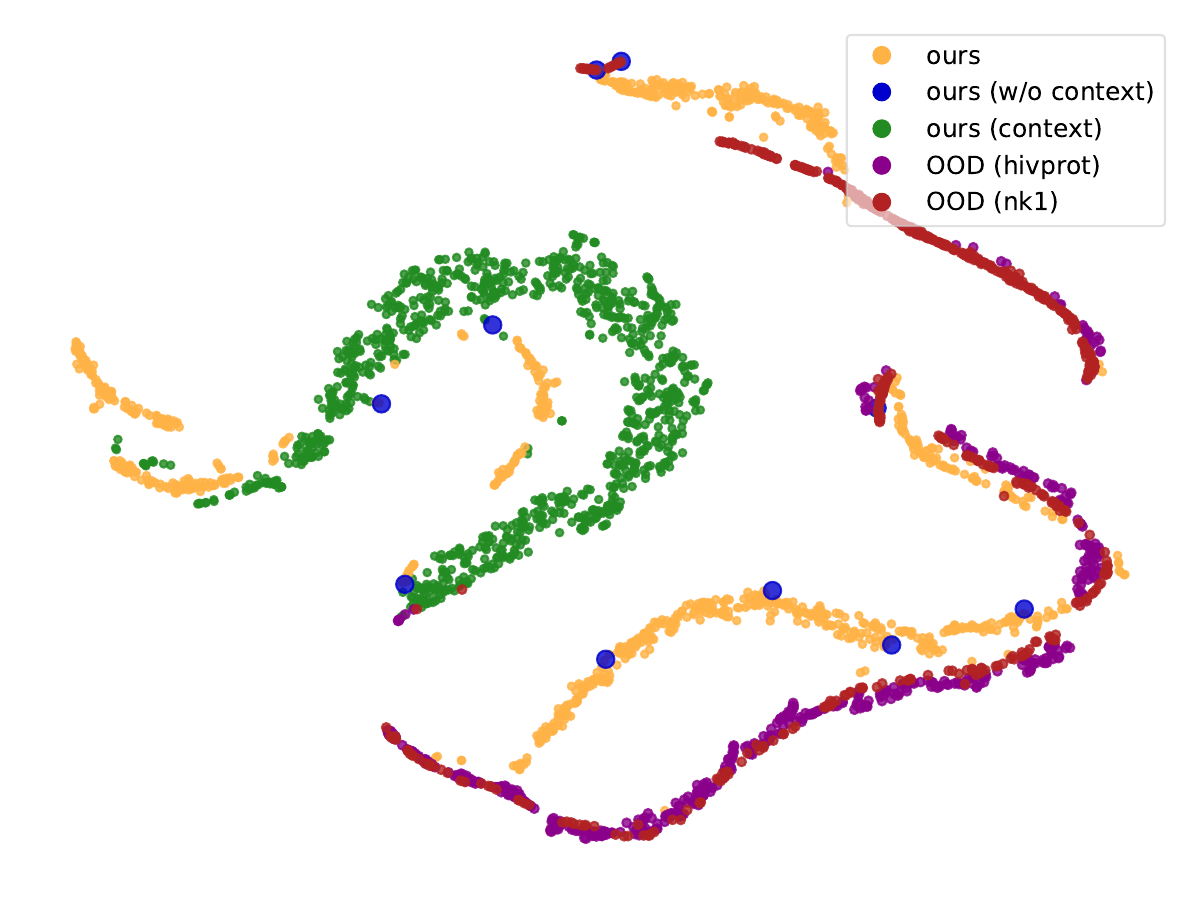}
    \caption{\textbf{Ours}}
\end{subfigure}
\hfill
\begin{subfigure}[b]{0.32\linewidth}
    \centering
    \includegraphics[width=\linewidth]{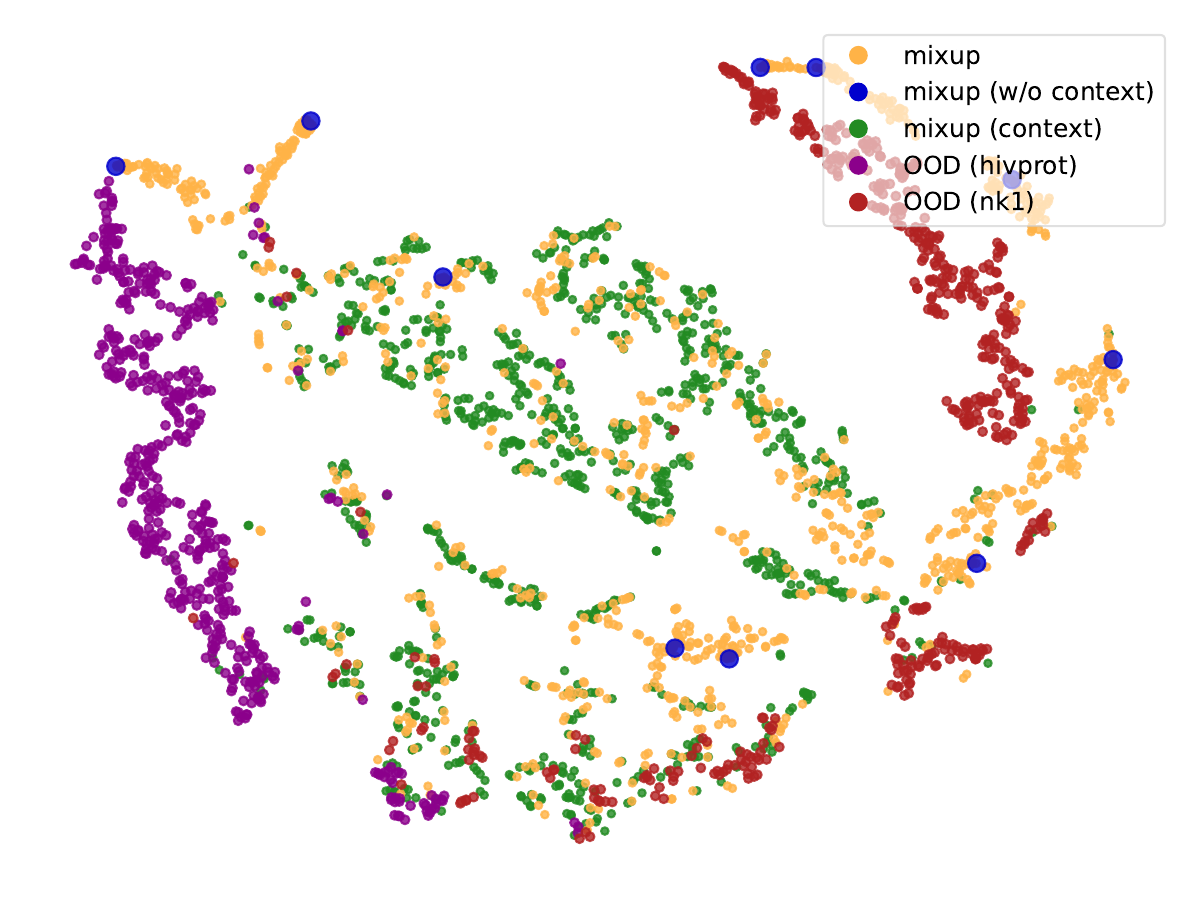}
    \caption{\textbf{Mixup (w/ outlier exposure)}}
\end{subfigure}
\hfill
\begin{subfigure}[b]{0.32\linewidth}
    \centering
    \includegraphics[width=\linewidth]{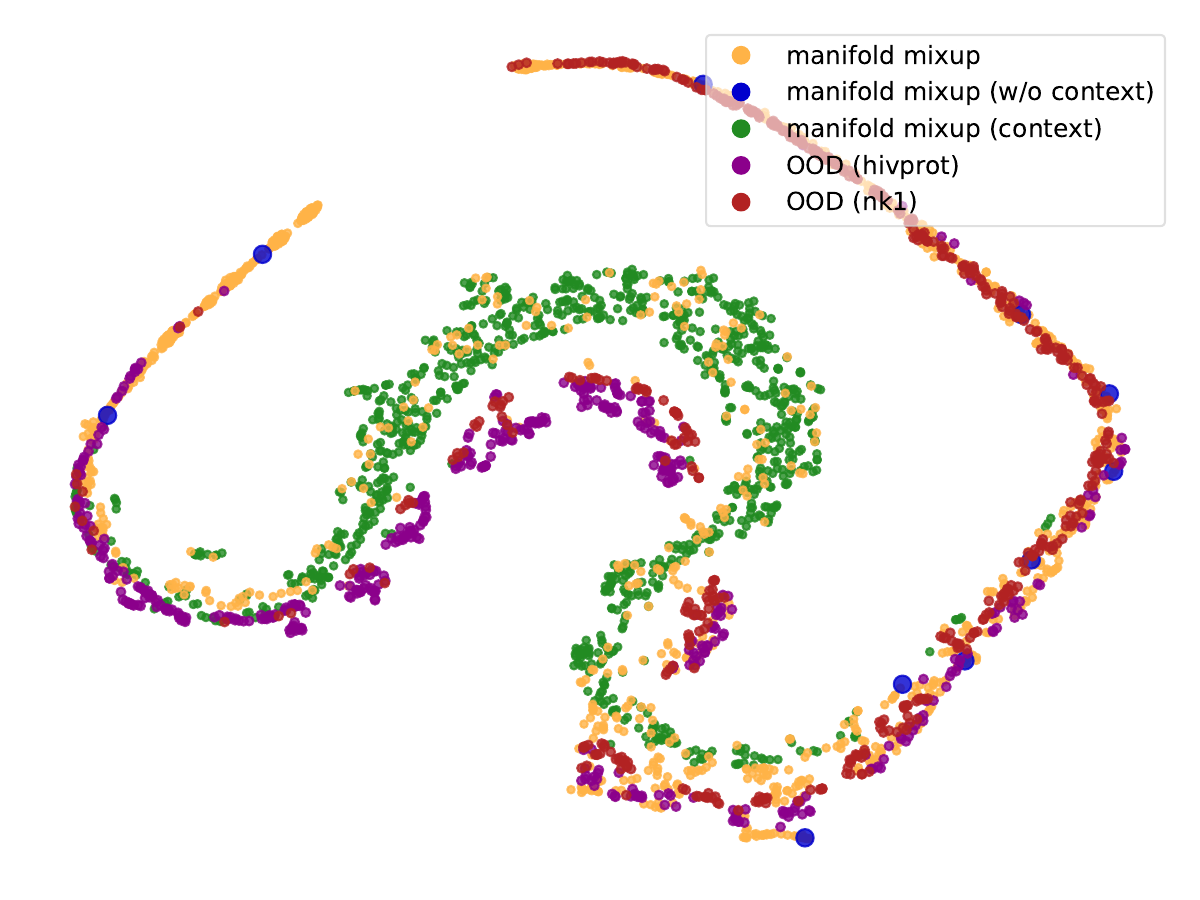}
    \caption{\textbf{Manifold Mixup (w/ bilevel optim.)}}
\end{subfigure}
\vspace{-0.5em}
\caption{\small \textbf{t-SNE visualization of the model trained on the DPP4 (count) dataset}}
\label{fig:app_tsne_dpp4_count}
\end{figure}

\begin{figure}[t]
\centering
\begin{subfigure}[b]{0.32\linewidth}
    \centering
    \includegraphics[width=\linewidth]{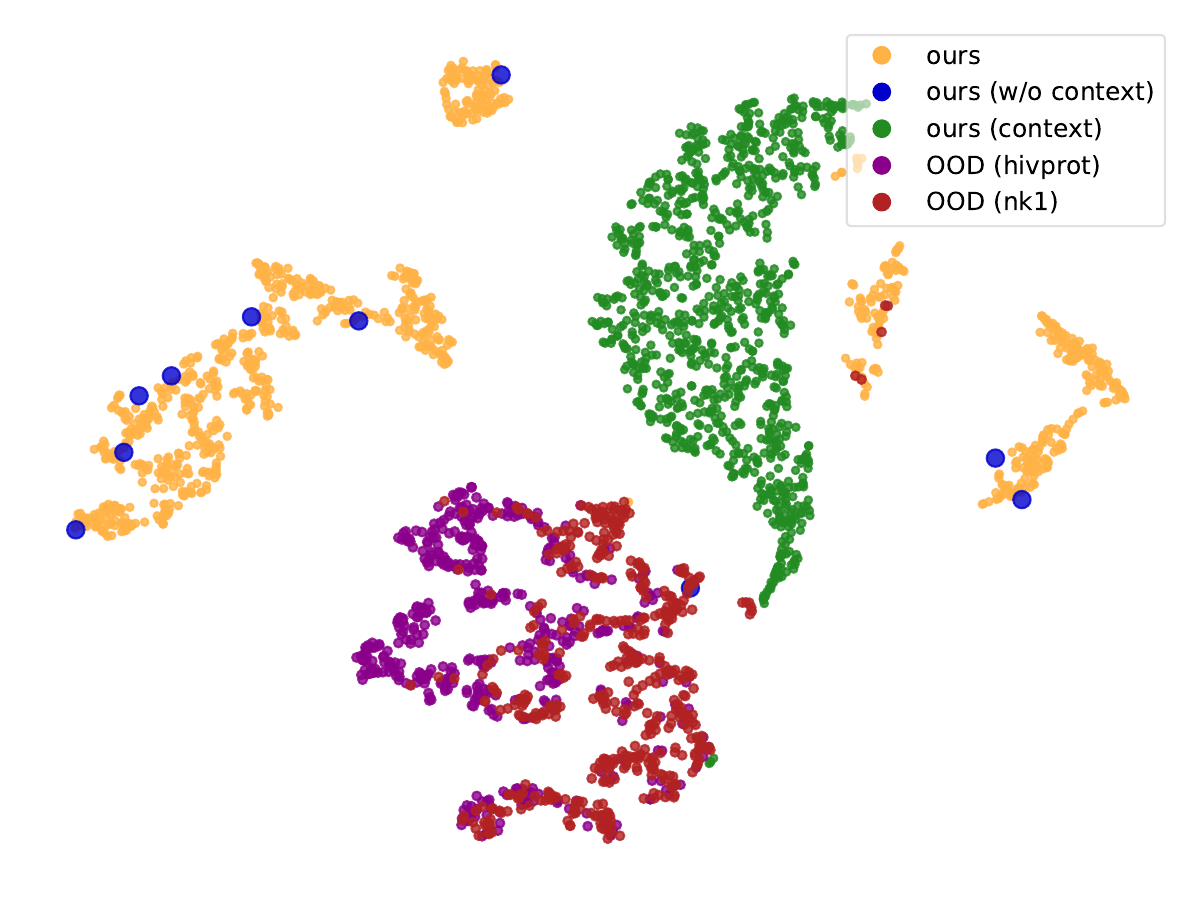}
    \caption{\textbf{Ours}}
\end{subfigure}
\hfill
\begin{subfigure}[b]{0.32\linewidth}
    \centering
    \includegraphics[width=\linewidth]{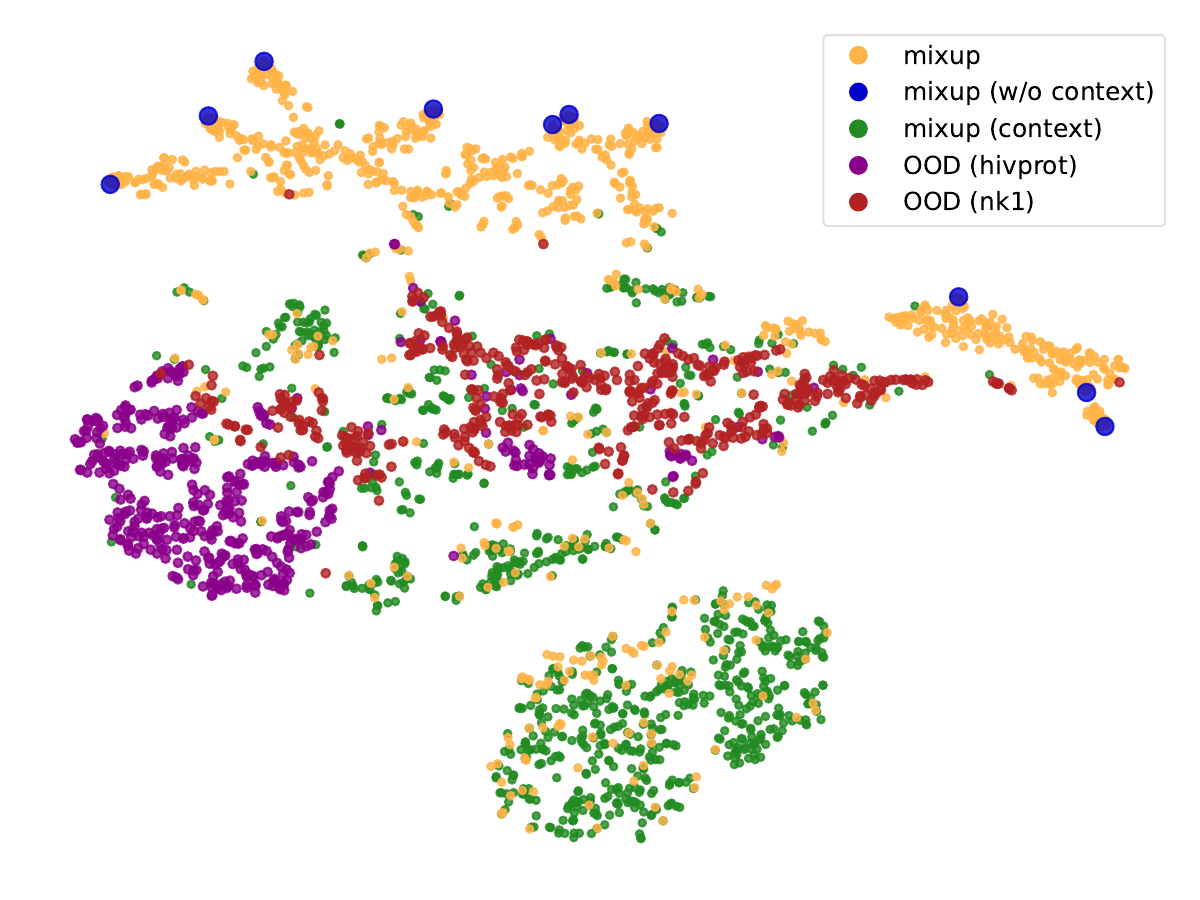}
    \caption{\textbf{Mixup (w/ outlier exposure)}}
\end{subfigure}
\hfill
\begin{subfigure}[b]{0.32\linewidth}
    \centering
    \includegraphics[width=\linewidth]{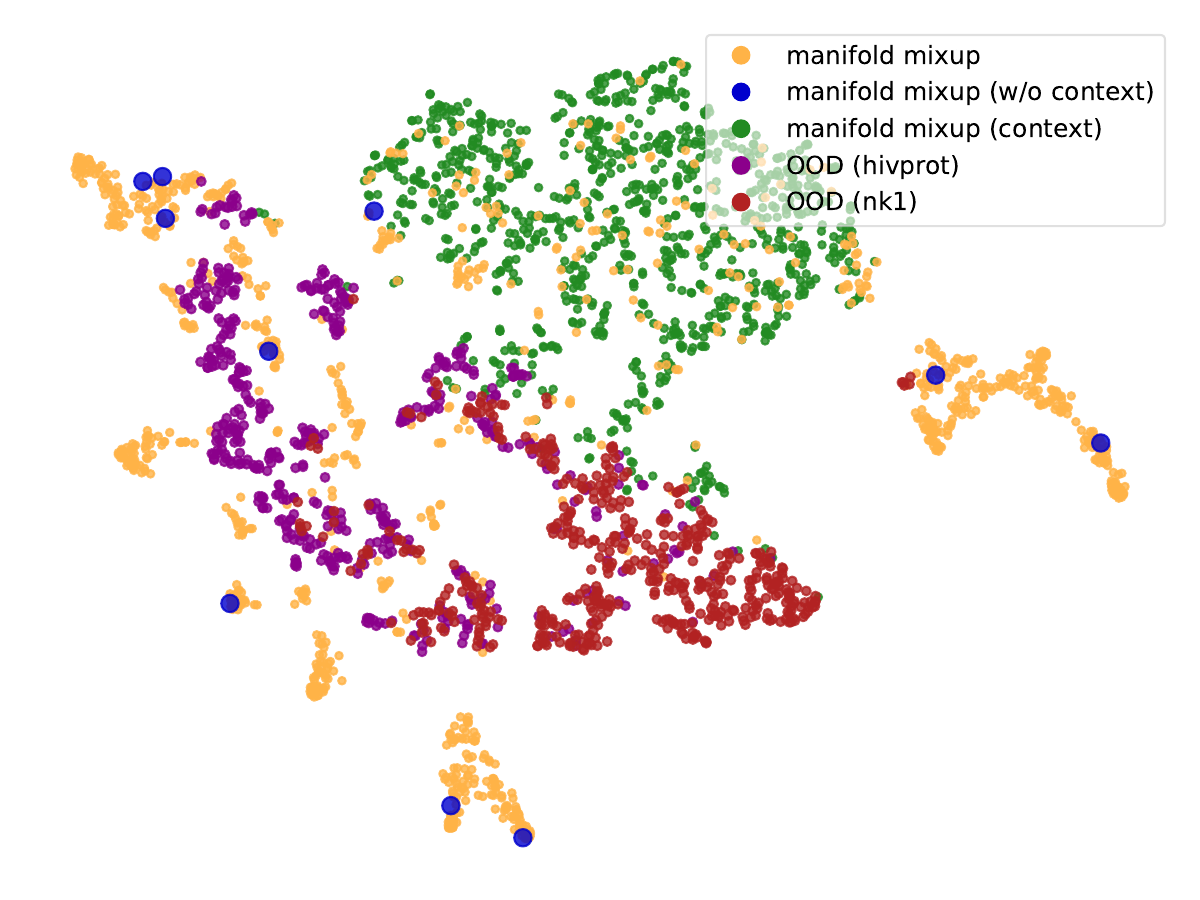}
    \caption{\textbf{Manifold Mixup (w/ bilevel optim.)}}
\end{subfigure}
\vspace{-0.5em}
\caption{\small \textbf{t-SNE visualization of the model trained on the DPP4 (bit) dataset}}
\label{fig:app_tsne_dpp4_bit}
\end{figure}

\begin{figure}[t]
\centering
\begin{subfigure}[b]{0.32\linewidth}
    \centering
    \includegraphics[width=\linewidth]{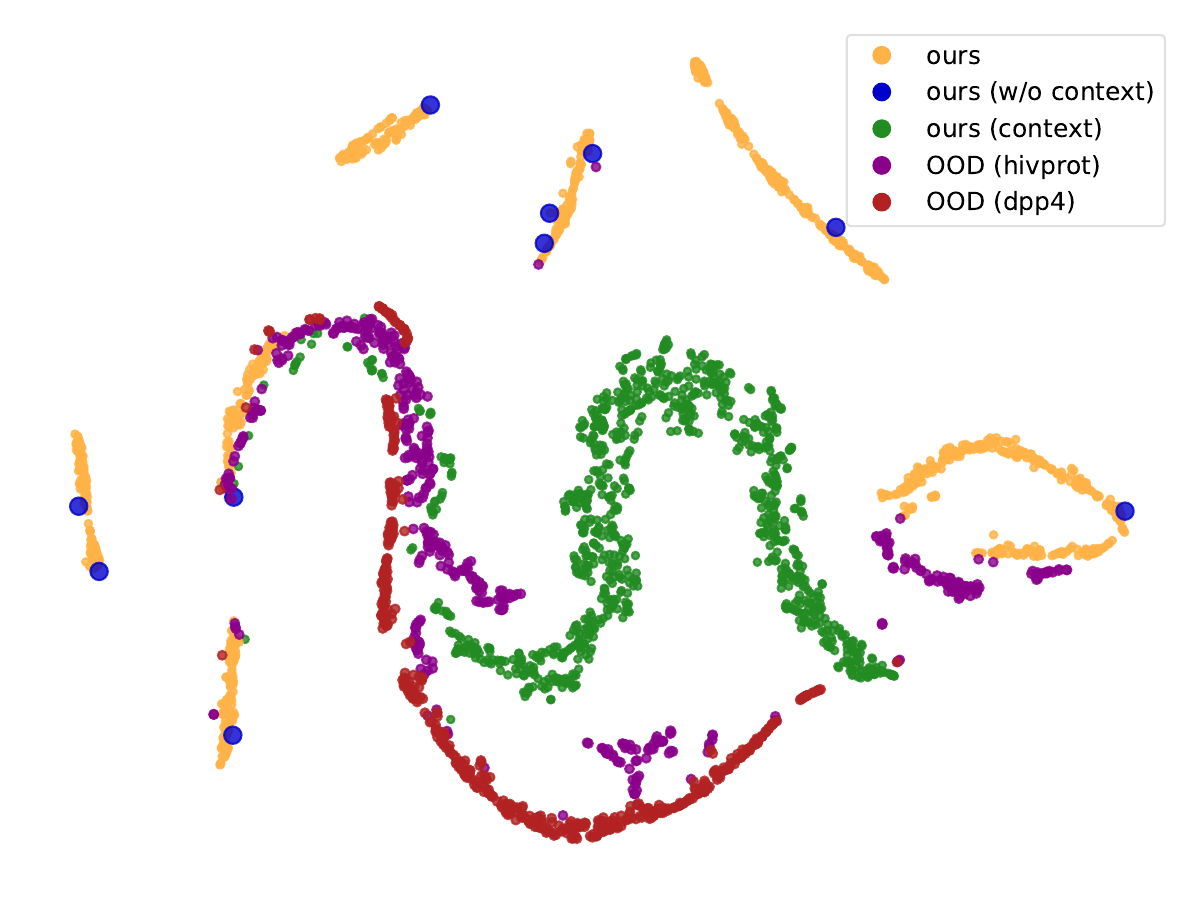}
    \caption{\textbf{Ours}}
\end{subfigure}
\hfill
\begin{subfigure}[b]{0.32\linewidth}
    \centering
    \includegraphics[width=\linewidth]{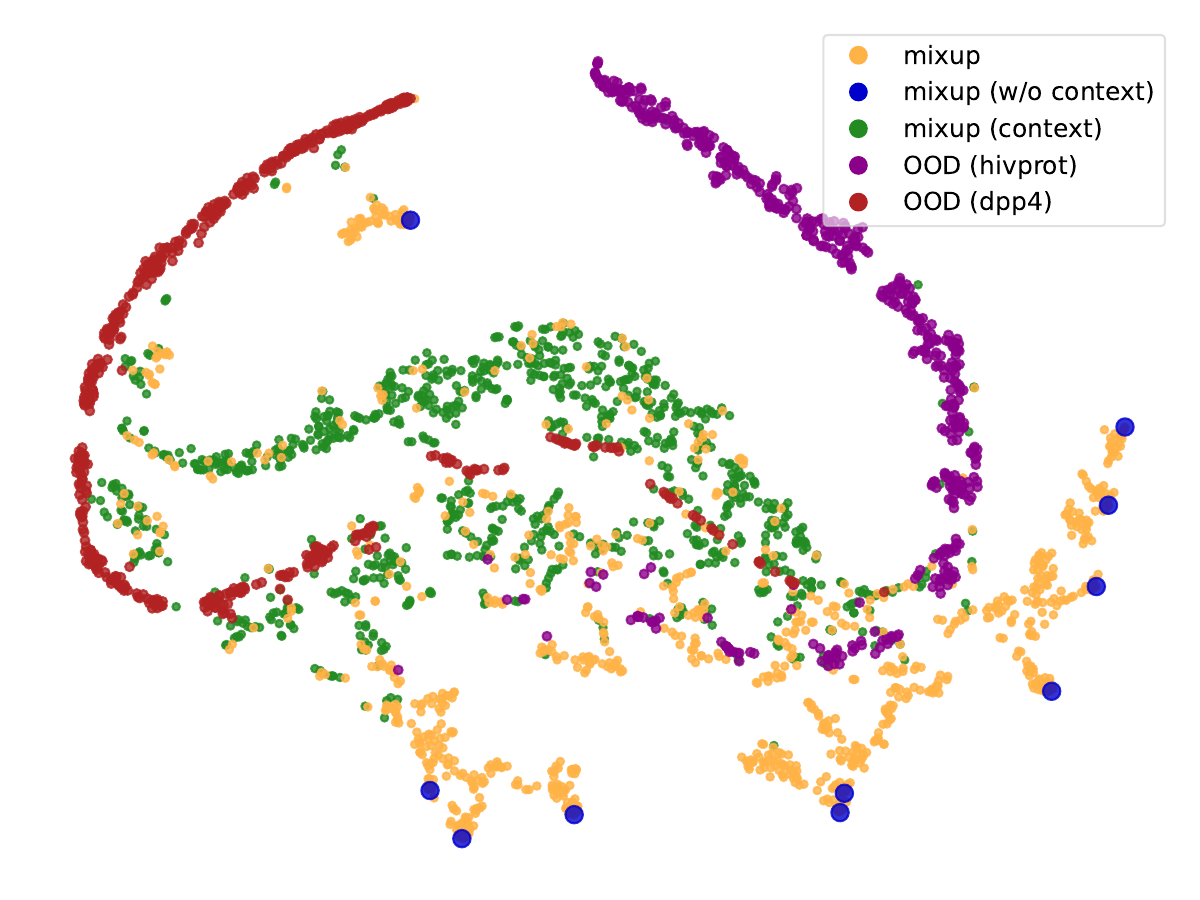}
    \caption{\textbf{Mixup (w/ outlier exposure)}}
\end{subfigure}
\hfill
\begin{subfigure}[b]{0.32\linewidth}
    \centering
    \includegraphics[width=\linewidth]{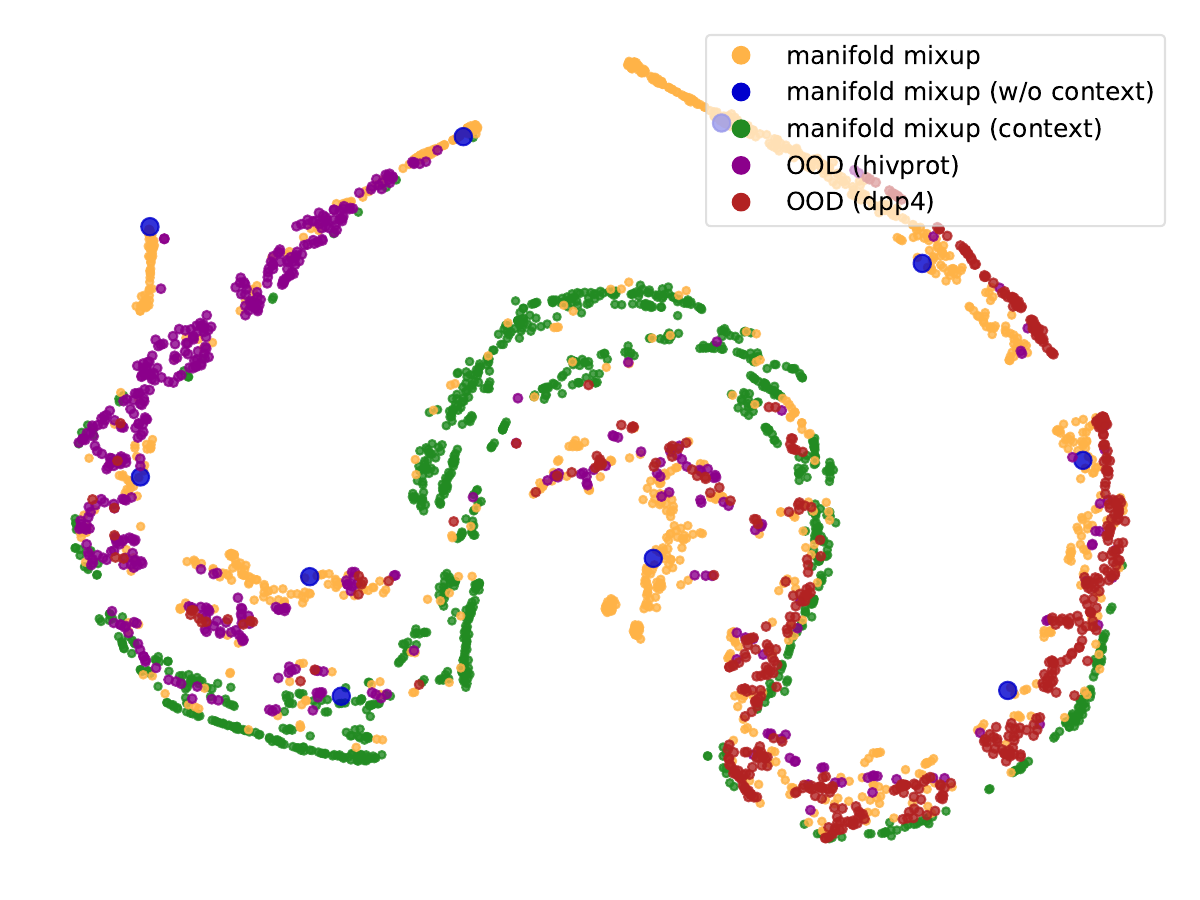}
    \caption{\textbf{Manifold Mixup (w/ bilevel optim.)}}
\end{subfigure}
\vspace{-0.5em}
\caption{\small \textbf{t-SNE visualization of the model trained on the NK1 (count) dataset}}
\label{fig:app_tsne_nk1_count}
\end{figure}

\begin{figure}[t]
\centering
\begin{subfigure}[b]{0.32\linewidth}
    \centering
    \includegraphics[width=\linewidth]{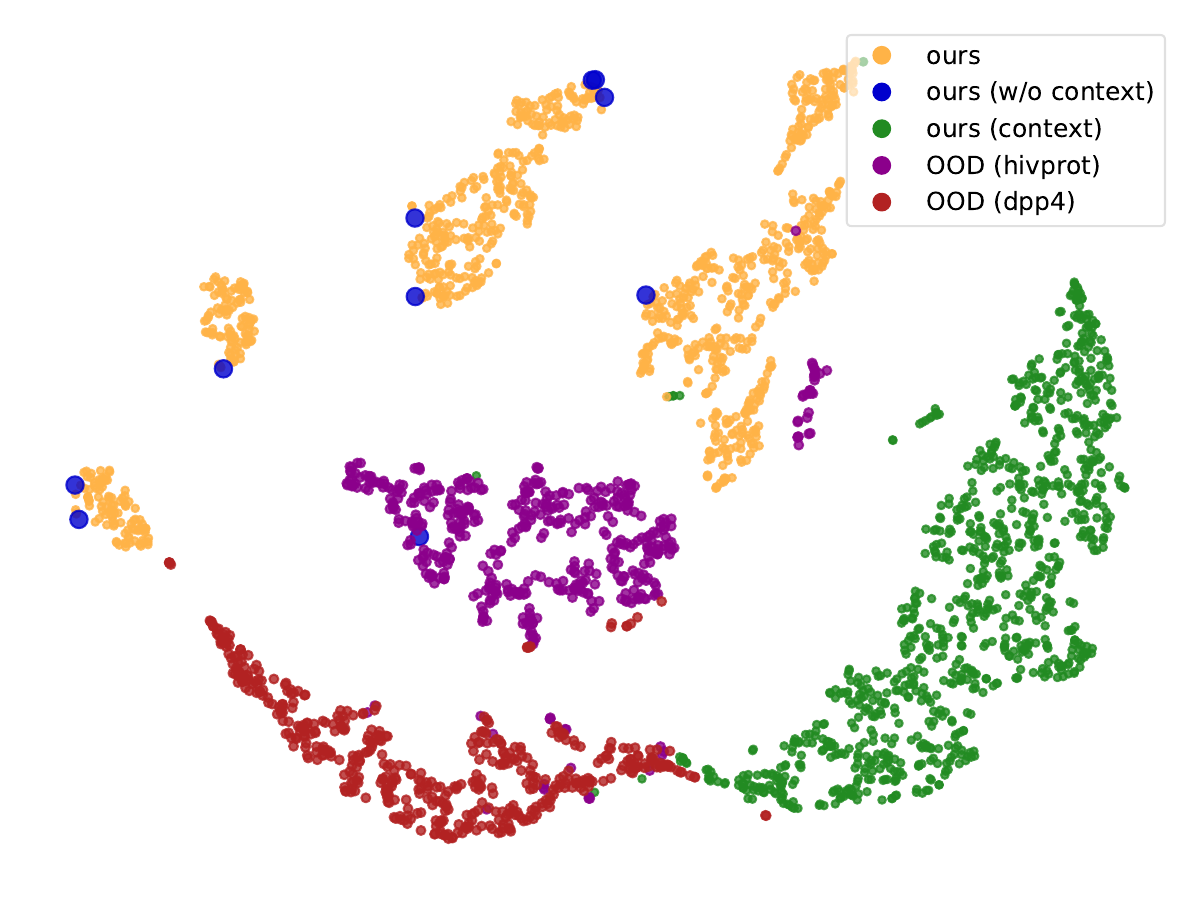}
    \caption{\textbf{Ours}}
\end{subfigure}
\hfill
\begin{subfigure}[b]{0.32\linewidth}
    \centering
    \includegraphics[width=\linewidth]{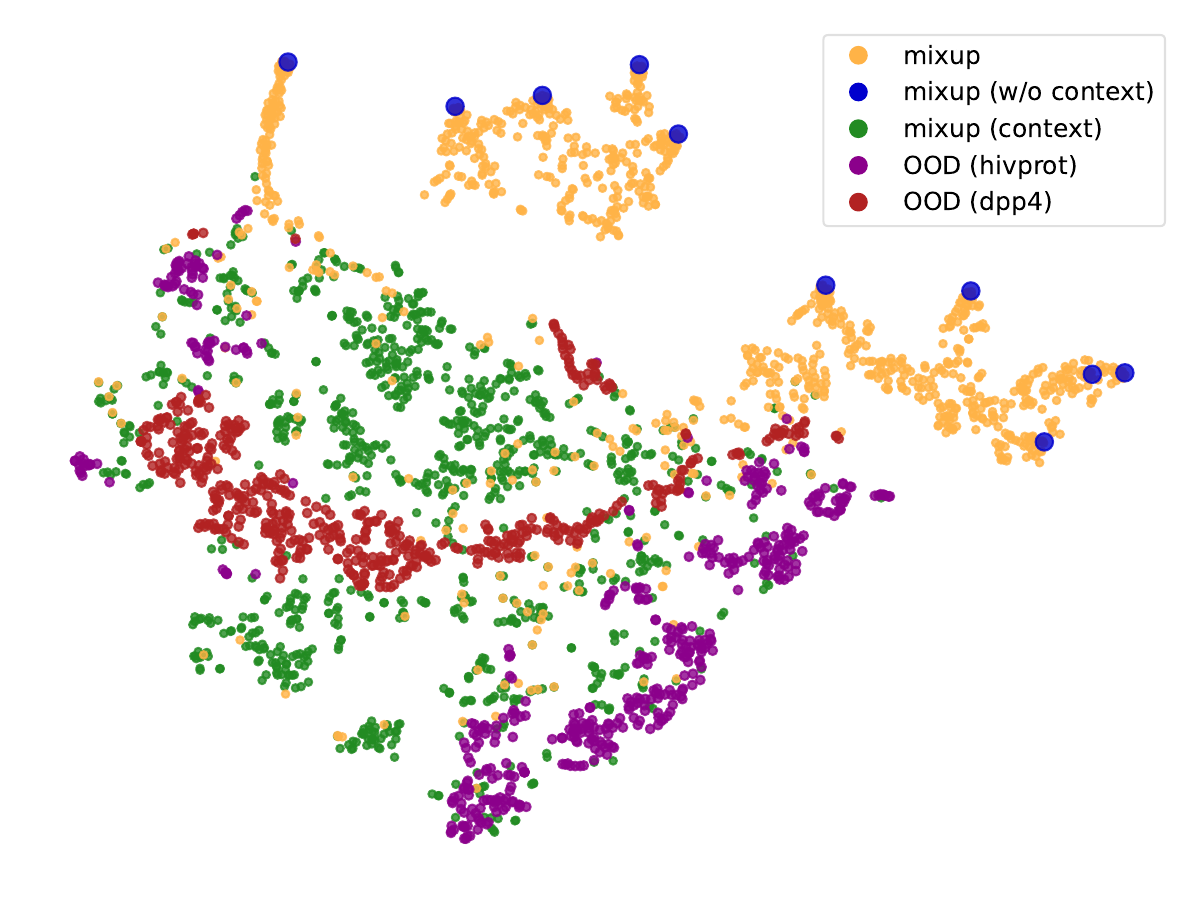}
    \caption{\textbf{Mixup (w/ outlier exposure)}}
\end{subfigure}
\hfill
\begin{subfigure}[b]{0.32\linewidth}
    \centering
    \includegraphics[width=\linewidth]{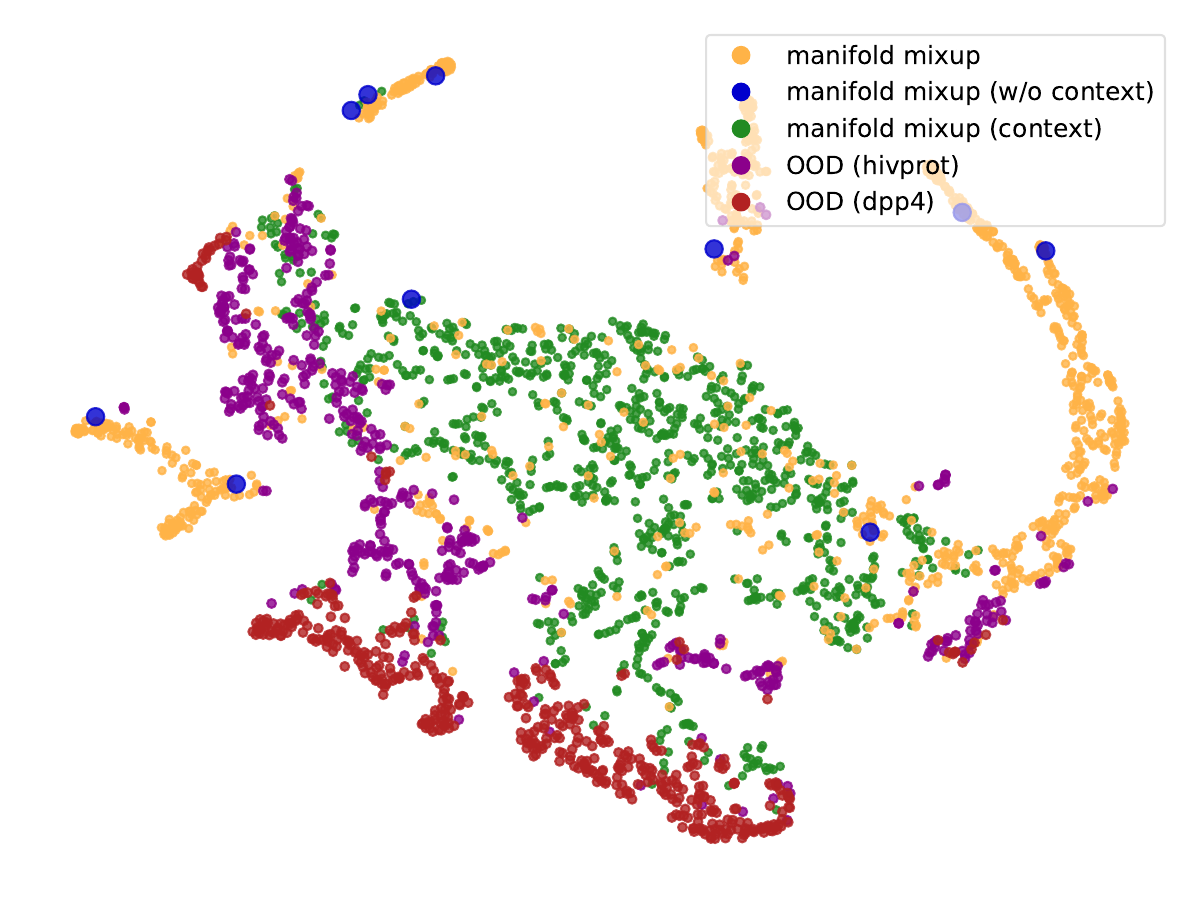}
    \caption{\textbf{Manifold Mixup (w/ bilevel optim.)}}
\end{subfigure}
\vspace{-0.5em}
\caption{\small \textbf{t-SNE visualization of the model trained on the NK1 (bit) dataset}}
\label{fig:app_tsne_nk1_bit}
\end{figure}

\begin{wraptable}[7]{r}{5.5cm}
\vspace{-2em}
    \caption{OOD datasets for t-SNE visualizations.}
    \label{tab:ood-tsne}
\resizebox{\linewidth}{!}{
  \begin{tabular}{ll}
    \toprule
     Train dataset & OOD Datasets \\
     \midrule
     HIVPROT & DPP4, NK1 \\ 
     DPP4 & HIVPROT, NK1 \\ 
     NK1 & DPP4, HIVPROT \\ 
     \bottomrule
\end{tabular}  
}
\end{wraptable}
In \cref{fig:tsne_nk1_bit}, we visualize the penultimate-layer embeddings \( \mathbf{z}_i = x_i^{(L-1)} \) using t-SNE~\citep{tsne}, across three models that interpolate \( \mathcal{D}_\text{train} \) with \( \mathcal{D}_\text{context} \). For each trained model, we generate the following embeddings: \( \mathbf{Z}_{\text{joint}} \): obtained by passing both \( x_i \) and \( C_i \) to the model; \( \mathbf{Z}_{\text{input}} \): obtained by passing only \( x_i \); \( \mathbf{Z}_{\text{context}} \): obtained by passing only \( C_i \). This allows us to observe how the model handles each type of input. To add an additional qualitative (OOD) evaluation, we utilize one of the Merck datasets~\citep{merck_activity} that was not used during training (see~\cref{tab:ood-tsne}). Using this OOD dataset, we generate the embeddings of \( \mathbf{Z}_{\text{ood}} \), by only passing \( x_i^{\text{ood}} \) to the model. In addition to~\cref{fig:tsne_nk1_bit}, we also provide extra examples in~\cref{fig:app_tsne_hivprot_count,fig:app_tsne_hivprot_bit,fig:app_tsne_dpp4_count,fig:app_tsne_dpp4_bit,fig:app_tsne_nk1_count,fig:app_tsne_nk1_bit}.

\subsection{t-SNE plots across all datasets}

Here, we present the t-SNE plots for all datasets used in our experiments, as listed in \cref{tab:merck_results}. For each in-distribution model, we visualize the corresponding two out-of-distribution datasets. For our method, among the DeepSets~\citep{deepsets} and Set Transformer~\citep{set-transformer} variants, we visualize the model that achieved the highest performance in \cref{tab:merck_results}.

\clearpage

\clearpage


\end{document}